\documentclass[11pt]{article}

\usepackage[preprint]{acl}

\usepackage{times}
\usepackage{latexsym}

\usepackage{colortbl}
\definecolor{MemoryForgeRow}{RGB}{240,244,250}
\usepackage[T1]{fontenc}

\usepackage[utf8]{inputenc}

\usepackage{microtype}

\usepackage{inconsolata}

\usepackage{graphicx}
\usepackage{amsmath}
\usepackage{amssymb}
\usepackage{booktabs}
\usepackage{multirow}
\usepackage{placeins}
\usepackage{subcaption}
\usepackage[most]{tcolorbox}

\definecolor{PromptFrame}{RGB}{86,103,128}
\definecolor{PromptBack}{RGB}{248,250,253}
\definecolor{PromptTitleBack}{RGB}{230,236,245}
\definecolor{PromptRoleBack}{RGB}{222,230,241}

\newtcbox{\PromptRoleBox}{%
  on line,
  arc=1pt,
  boxrule=0.25pt,
  colback=PromptRoleBack,
  colframe=PromptFrame,
  left=1.5pt,
  right=1.5pt,
  top=0pt,
  bottom=0pt,
  boxsep=1pt
}
\newcommand{\PromptRole}[1]{\par\smallskip\noindent\PromptRoleBox{\texttt{#1}}\quad}
\newtcolorbox{PromptCard}[1]{%
  enhanced,
  breakable,
  colback=PromptBack,
  colframe=PromptFrame,
  colbacktitle=PromptTitleBack,
  coltitle=black,
  fonttitle=\bfseries\small,
  title={#1},
  boxrule=0.45pt,
  arc=2pt,
  left=4pt,
  right=4pt,
  top=3pt,
  bottom=3pt,
  before skip=5pt,
  after skip=5pt,
  borderline west={1.2pt}{0pt}{PromptFrame}
}

\newcommand{\tightdisplayspacing}{%
  \setlength{\abovedisplayskip}{2pt}%
  \setlength{\belowdisplayskip}{2pt}%
}
\AtBeginEnvironment{equation}{\tightdisplayspacing}
\AtBeginEnvironment{equation*}{\tightdisplayspacing}
\AtBeginEnvironment{align}{\tightdisplayspacing}
\AtBeginEnvironment{align*}{\tightdisplayspacing}

\title{MemoryForge: Synthesize Lifelong Memory for Human-Like LLM Agents}

\author{Bohan Tang \\
  LIGHTSPEED \\
  \texttt{bohantang@global.tencent.com} \And
  Yiwen Guo\textsuperscript{\textdagger} \\
  Independent Researcher \\
  \texttt{guoyiwen89@gmail.com} }

\begin{document}
\maketitle
\begin{abstract}
Equipping Large Language Models (LLMs) with human-like personas is crucial for agentic applications, such as role-play and user simulation. Traditional prompt-based methods rely on \textbf{descriptive conditioning} by injecting static textual profiles, which often makes agents show generic behaviors due to a lack of realistic life memory. To fill this gap, we introduce \textbf{memory-based conditioning}, a paradigm inspired by the cognitive psychology, which replaces abstract profiles with an autobiographical memory base, enabling frozen LLMs to dynamically retrieve situation-relevant memory to guide their behaviors. We formalize its enabling task as \textbf{customized lifelong memory synthesis} and propose \textbf{MemoryForge}, a novel framework to synthesize such lifelong memory from brief target personas. MemoryForge has three key components: a context generator for socio-historical grounding, a life organizer for developmental coherence toward the target identity, and a multi-resolution simulator that balances broad temporal summaries with high-fidelity episodic experiences. Experiments on PersonaGym for role-play and SimulatorArena for user-simulation, show that the synthesized memory base by MemoryForge enables frozen LLMs to exhibit more human-like behaviors than strong descriptive conditioning baselines across multiple metrics and LLM backbones.  Code is available \href{https://github.com/Tencent/MemoryForge}{\underline{here}}.
\end{abstract}
\let\thefootnote\relax\footnotetext{\textsuperscript{\textdagger} Corresponding author.}

\section{Introduction}

As Large Language Models (LLMs) drive the widespread deployment of AI agents \citep{sumers2024cognitive, wang2024survey, xi2025rise}, emerging applications increasingly demand that such systems move beyond task execution to exhibit human-like behaviors grounded in specific personas. For instance, role-play agents must maintain identities in open-ended dialogues \citep{park2023generative, charactereval}, while effective user simulators must reflect human dynamics \citep{dou2025simulatorarena, sim2realgap}. Hence, making an LLM act consistently with a target persona is increasingly important.

To achieve this, fine-tuning methods embed persona into model parameters with extensive dialogue data \citep{shao2023characterllm, wang2024rolellm, yang-etal-2025-crafting}. While effective, per-persona training is costly and inapplicable to proprietary APIs, limiting its practicality. These constraints lead to interest in prompt-based methods under a paradigm we term \textbf{descriptive conditioning}. Keeping the backbone LLM frozen, this paradigm injects static textual profiles, such as demographic-anchored backstories \citep{moon2024virtual} and taxonomy-guided narratives \citep{wang2025deeppersona}, into the context window to guide downstream LLM behaviors.

\begin{figure*}[t]
  \centering
  \includegraphics[width=1.9\columnwidth]{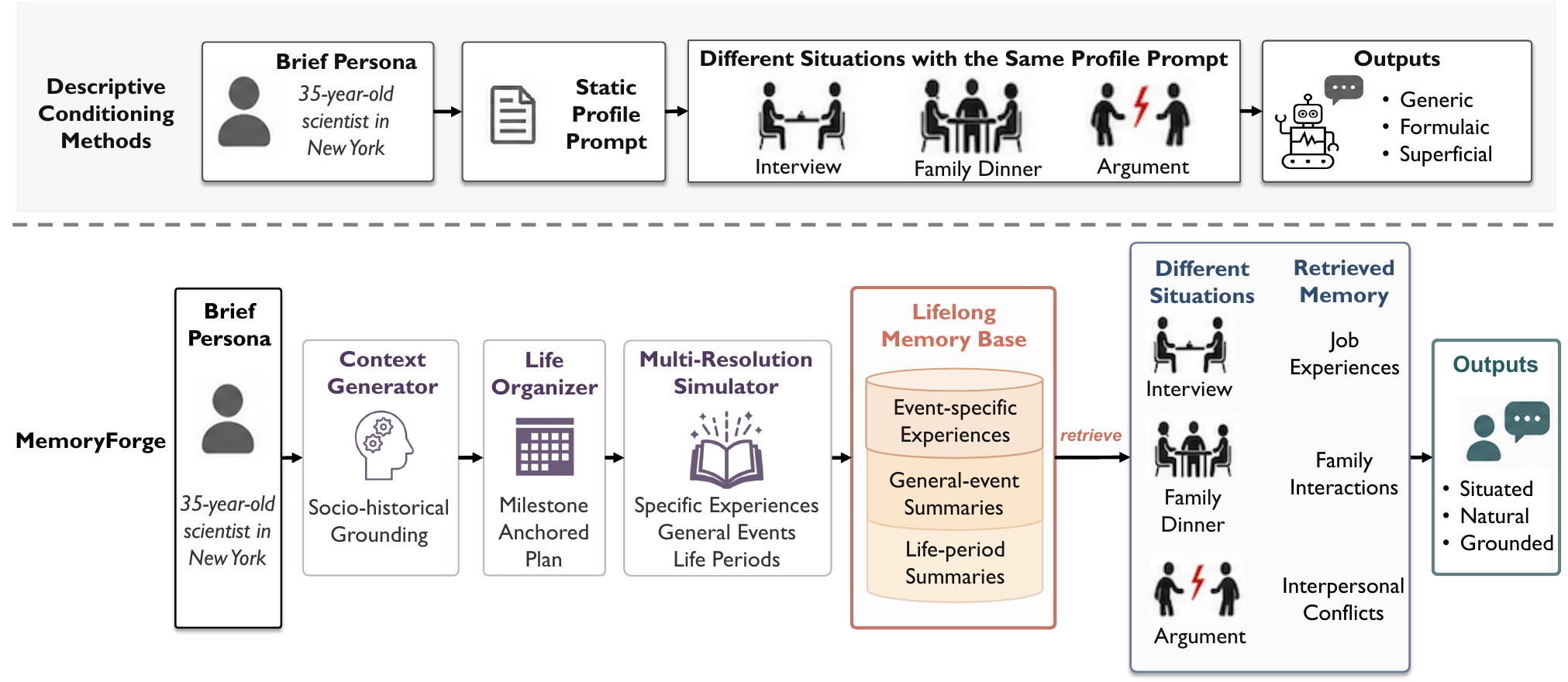}
  \vspace{-2mm}
  \caption{Comparison between existing descriptive conditioning methods and MemoryForge.}
  \label{fig:memoryforge-vs-flat}
\vspace{-3mm}
\end{figure*}

However, according to cognitive psychology, descriptive conditioning suffers from two key limitations. First, \textit{static injection}: the persona text is typically fed verbatim to every interaction \citep{ge2024scaling,moon2024virtual}. Whether the agent faces an interview or a family dinner, the persona text remains identical. This conflicts with how human cognition operates, which activates different memory based on current situations \citep{conway2005, tulving1983elements}. Second, \textit{behavioral under-specification}: the persona is usually grounded in trait labels and scripted dialogues \citep{wang2025deeppersona}, whereas psychology shows that behavioral dispositions are better understood not merely as abstract descriptors, but as patterns shaped by accumulated experiences across the life \citep{erikson1963childhood, costa1999five}. For instance, a ``cautious'' individual defrauded by a friend acts differently than one raised in a risk-averse household, as the underlying causal memory differs. Lacking situation-relevant grounding in life memory, models default to pretrained stereotypes, producing homogenized behavior \citep{sim2realgap}. Hence, addressing both limitations requires a memory base that (1) enables situation-relevant retrieval, and (2) spans a comprehensive developmental trajectory.

In this work, we explore a novel paradigm: \textbf{memory-based conditioning}. Rather than injecting a fixed profile that \textit{describes} the persona, we synthesize a structured lifelong memory base from which the frozen LLM \textit{retrieves} situation-relevant memory as dynamic conditioning context.

Following this spirit, we formalize the enabling task as \textbf{customized lifelong memory synthesis}: given a brief persona description, produce a memory base spanning the full developmental trajectory. This task demands jointly satisfying three desiderata: the trajectory must be \textit{realistic} in its socio-historical grounding, \textit{controllable} toward the target identity, and \textit{efficient} across a decades-long lifespan.  While simulating a character's life is a natural approach, existing simulators \citep{park2023generative, piao2025agentsociety, zhang2025socioverse, duan2026lifesimlonghorizonuserlife} fall short on all three. Their environments are usually fictional sandboxes or present-day snapshots, lacking year-by-year grounding. As open-ended forward engines, they offer limited mechanisms to steer toward a target identity. Further, they usually simulate at a single temporal resolution, making a full lifespan computationally prohibitive.

To fill these gaps, we propose \textbf{MemoryForge}, a novel framework that realizes customized lifelong memory synthesis with three key LLM-driven components. A Context Generator achieves \textit{realism} by inferring plausible socio-cultural and historical context from the description. A Life Organizer achieves \textit{controllability} by decomposing the lifespan into milestone-anchored periods that ensure convergence toward the target identity. A Multi-Resolution Simulator achieves \textit{efficiency} by mirroring the three-level hierarchy of human autobiographical memory. It generates life-period summaries for routine phases, general-event summaries for recurring activities, and high-fidelity event-specific experiences for identity-shaping moments, making decades tractable while producing a memory structure naturally compatible with cognitively motivated retrieval. Figure~\ref{fig:memoryforge-vs-flat} shows the contrast with existing methods.

We conduct extensive experiments on two public benchmarks that capture orthogonal forms of human-likeness. PersonaGym measures whether an agent remains human-like to a persona in role-play \citep{personagym}. SimulatorArena measures whether an agent behaves like real users in task-oriented settings \citep{dou2025simulatorarena}. On both benchmarks, MemoryForge outperforms 5 strong baselines \citep{moon2024virtual,ge2024scaling,liu2025big5chat,wang2025deeppersona, yang-etal-2025-crafting} in multiple metrics under 2 different LLM backbones. Our contributions are fourfold:

$\bullet$ We propose \textbf{memory-based conditioning} as a novel persona-conditioning paradigm, where LLMs are guided by situation-relevant memory.

$\bullet$ We formalize \textbf{customized lifelong memory synthesis}, the task of generating a lifelong autobiographical memory base from a brief target persona.


$\bullet$ We propose \textbf{MemoryForge}, a novel framework that synthesizes lifelong autobiographical memory bases for human-like LLM agents.

$\bullet$ We empirically show that the synthesized lifelong memory improves both role-play and user-simulation human-likeness for frozen LLMs.

\section{Related Works}
\label{sec:related}
\textbf{LLM agent persona.}
Equipping LLMs with target personas is studied along two tracks. \emph{Fine-tuning} bakes a character into the weights using persona-specific corpora and trait-labelled data~\citep{shao2023characterllm,wang2024rolellm,zhou2024characterglm,yang-etal-2025-crafting, liu2025big5chat}, but it is computationally expensive, inapplicable to the closed-source APIs that dominate real-world scenarios, and impractical when personas must be frequently revised. \emph{Prompt-based} methods keep the backbone LLM frozen and engineer the input context. These methods usually fall under what we term the \textit{descriptive conditioning} paradigm, spanning a spectrum of descriptive detail: one-paragraph flat cards~\citep{ge2024scaling}, demographically anchored backstories~\citep{moon2024virtual}, and taxonomy-sampled deep narratives with hundreds of attributes~\citep{wang2025deeppersona}. Despite this progression toward richer descriptions, prompt-based methods share a common format: a fixed text without situation-relevant grounding in lived memory injected into the context window. Our \textit{memory-based conditioning} paradigm departs from this descriptive axis. It equips the frozen LLM with a structured lifelong autobiographical memory base, enabling dynamic, situation-relevant memory retrieval on demand.

\textbf{LLM-based simulation.}
LLM-based simulation has progressed rapidly along two complementary axes. \emph{Society-level} simulators populate shared worlds with many agents to study emergent collective behavior \citep{park2023generative,piao2025agentsociety,zhang2025socioverse}. \emph{Individual-level} simulators target the behavioral realism of one user's actions, conditioned on a profile already in hand, whether derived from a brief description \citep{wang2025simulating}, extracted from interaction logs \citep{duan2026lifesimlonghorizonuserlife}, or reconstructed from deep interviews of real people \citep{park2026llmagentsgroundedselfreports}. The shared paradigm is \emph{forward projection}: given a starting profile, simulate what happens next. Conversely, MemoryForge inverts this direction: given a target identity, synthesize the lifelong memory base from which that identity would have emerged. Thereon, we reframe simulation as a controlled generative primitive for memory synthesis.

\textbf{Memory for LLM agents.}
Memory has emerged as an important object for LLM agents along three lines: organising interaction history into structured stores~\citep{xu2026amem,kang-etal-2025-memory,liu2026simplemem}, carving out a persistent self-substrate that resists drift~\citep{platnick2025id,He2025HelloAgain}, and learning when and what to write, retrieve, or forget~\citep{cai2026askneededproactiveretrieval,zhang2026deltamem}. Across this literature the input is assumed to \emph{already exist} from past dialogues or observed trajectories, none of them produces a memory base for a \emph{fresh} virtual persona who has never been observed. Rather than \emph{managing} memory that arrives from elsewhere, we \emph{synthesize} a lifelong autobiographical memory base from a brief persona description.

\section{Methodology}
\label{sec:method}
This section starts by formalizing the \textbf{customized lifelong memory synthesis} task (\S\ref{sec:problem}). Thereon, we elaborates on the three key components of our \textbf{MemoryForge}, a novel framework that autonomously synthesizes lifelong autobiographical memory from brief persona descriptions (\S\ref{sec:memoryforge}).

\subsection{Problem Formulation}
\label{sec:problem}
Given a brief persona description $\pi$ (e.g., ``a 35-year-old scientist in New York''), our goal is to synthesize an autobiographical memory base,
\begin{equation*}
\mathcal{M}_\pi \;=\; (\mathcal{L}, \mathcal{G}, \mathcal{E}),
\end{equation*}
following the hierarchy of human autobiographical memory~\citep{conway2000sms}. The three levels are life periods $\mathcal{L}=\{\ell_i\}_{i=1}^{P}$, general events $\mathcal{G}=\{g_j\}_{j=1}^{N_g}$, and event-specific experiences $\mathcal{E}=\{e_k\}_{k=1}^{N_e}$: each $\ell_i$ summarises a semantically stable life phase, each $g_j$ captures recurring or typical events, and each $e_k$ records a concrete moment. Because $\pi$ specifies the target identity but leaves its developmental trajectory latent, we model synthesis as sampling from $p(\mathcal{M}_\pi \mid \pi)$ and require the resulting life history to be coherent and identity-consistent. Inspired by sociocultural theories of autobiographical memory~\citep{conway2000sms,nelson2004}, we operationalise this requirement through three desiderata: (1) \textit{realism}, which grounds memory in plausible socio-historical conditions; (2) \textit{controllability}, which keeps the trajectory self-coherent and ensures it entails $\pi$; and (3) \textit{efficiency}, which mirrors hierarchical human autobiographical memory rather than simulating decades uniformly.

\subsection{MemoryForge}
\label{sec:memoryforge}
To satisfy the aforementioned desiderata, we propose MemoryForge. We formalize the synthesis process of MemoryForge as a joint distribution to conceptually define its generation framework:
\begin{align}
p\left(\mathcal{M}_\pi \mid \pi\right)
\;=\;& \;\;\;\;\,p_{\mathrm{ctx}}\!\left(c \mid \pi\right) \cdot\; p_{\mathrm{org}}\!\left(\mathcal{P} \mid c, \pi\right) \nonumber \\
     & \cdot\!\prod_{i=1}^{P} p_{\mathrm{sim}}\!\left(\ell_i, \mathcal{G}_i,\mathcal{E}_i\bigl| \mathcal{P}_{\le i}, c, \pi, s_i\right),
\label{eq:factorisation}
\end{align}
where $c$ is a context containing basic persona attributes alongside the social and historical environment, $\mathcal{P}=\{\mathcal{P}_1,\dots,\mathcal{P}_P\}$ is a partition of the lifespan into life periods, and $\{\mathcal{G}_i,\mathcal{E}_i\}$ denote the period-local general and event-specific layers, and $s_i = \{\mathcal{M}_{<i}, \pi^{\mathrm{run}}_{<i}\}$ denote the accumulated memory base and running persona state before period $i$. We align the three factors in Eq.~\eqref{eq:factorisation} with the three modules of MemoryForge: the Context Generator models $p_{\mathrm{ctx}}$, the Life Organizer models $p_{\mathrm{org}}$, and the Multi-Resolution Simulator models $\prod_{i=1}^{P} p_{\mathrm{sim}}$. The detailed designs are in the following sections.


\begin{figure*}[t]
  \centering
  \includegraphics[width=1.8\columnwidth]{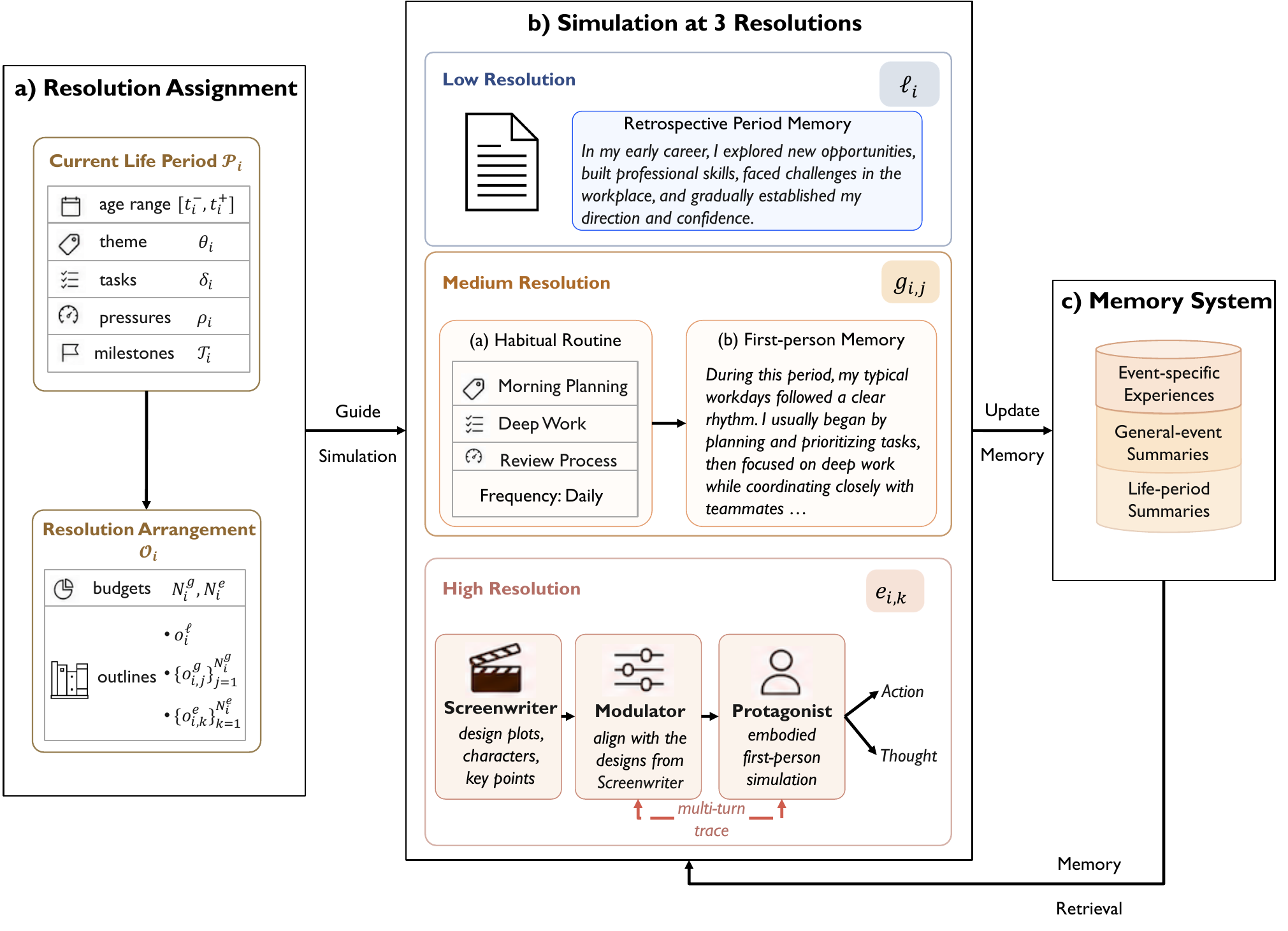}
  \vspace{-1mm}
  \caption{Overview of the Multi-Resolution Simulator in MemoryForge.} 
  \label{fig:multi-resolution-framework}
  \vspace{-1mm}
\end{figure*}

\subsubsection{Context Generator}
\label{sec:context-gen}

To ensure \textit{realism}, the Context Generator $p_{\mathrm{ctx}}(c\mid\pi)$ lifts the brief description $\pi$ into a comprehensive simulation seed: $c \;=\; (\phi, \sigma, \mathcal{R}),$
where $\phi$ denotes basic persona anchors, $\sigma$ is the surrounding social environment, and $\mathcal{R}$ is an initial social network. The generator comprises two steps.

\textbf{Personal context generation.} The first step uses an LLM to infer essential persona anchors from the brief description $\pi$. The resulting personal context $\phi$ includes demographic anchors, temporal anchors, growing-up and current locations, linguistic background, education and occupation targets, and identity-defining anchors extracted from $\pi$. 

\textbf{Social context generation.} Given the personal context $\phi$, the second step grounds the persona in a time-varying external world. For each year $t \in [t_{\mathrm{start}}, t_{\mathrm{end}}]$, an LLM retrieves and summarises the relevant real-world cultural, historical, and occupational background, yielding $\sigma \;=\; \{\sigma_t\}_{t=t_{\mathrm{start}}}^{t_{\mathrm{end}}}.$ Conditioned on both $\phi$ and the world timeline $\sigma$, it then instantiates the persona initial social network, $\mathcal{R} \;=\; (V, E, \{a_v\}_{v\in V}),$ where $V$ is the cast of supporting characters, $E$ specifies their ties to the protagonist, and $a_v$ records each character's basic profile. The resulting context $c=(\phi,\sigma,\mathcal{R})$ is passed to the subsequent modules.

\subsubsection{Life Organizer}
\label{sec:life-organizer}

To ensure \textit{controllability}, the Life Organizer $p_{\mathrm{org}}(\mathcal{P}\mid c,\pi)$ decomposes the lifespan $[t_{\mathrm{start}},t_{\mathrm{end}}]$ into developmentally coherent periods. 

\textbf{Life milestone generation.} Given $(\pi, c)$, an LLM first produces a set of trajectory milestones 
\begin{equation*}
\mathcal{T} \;=\; \{(t_m, \mu_m)\}_{m=1}^{M},
\end{equation*}
where each $\mu_m$ is a landmark event whose occurrence at calendar year $t_m$ is necessary for the final target identity $\pi$ to be plausible. By anchoring the trajectory in these milestones, this module ensures that all synthesized experiences tightly coordinate to entail the initial brief description, effectively preventing simulation drift.

\textbf{Life period generation.} With $(\pi, c, \mathcal{T})$, a second LLM call partitions the lifespan into a sequence of $P$ contiguous life periods:
\begin{align*}
\mathcal{P} &\;=\; (\mathcal{P}_1,\dots,\mathcal{P}_P), \nonumber \\
\mathcal{P}_i &\;=\; ([t_i^{\,-},t_i^{\,+}], \theta_i, \delta_i, \rho_i, \mathcal{T}_i),
\end{align*}
where each $\mathcal{P}_i$ comprises the age range $[t_i^{\,-},t_i^{\,+}]$, dominant theme $\theta_i$, developmental tasks $\delta_i$, salient pressures $\rho_i$, and the subset of milestones $\mathcal{T}_i \subset \mathcal{T}$ occurring within that span. Grounded in established life-span psychology \citep{erikson1963childhood,schwartz2012overview}, this partition ensures life experiences and major decisions naturally align with recognisable developmental phases. The resulting life trajectory $\mathcal{P}$ with the assembled context $c$ and the brief persona description $\pi$ are passed to the Multi-Resolution Simulator.

\subsubsection{Multi-Resolution Simulator}
\label{sec:simulator}

To efficiently realise $\prod_i p_{\mathrm{sim}}(\ell_i,\mathcal{G}_i,\mathcal{E}_i\mid \mathcal{P}_{\le i}, c, \pi, s_i)$, the Multi-Resolution Simulator mirrors the abstraction hierarchy of human memory~\citep{conway2000sms}. Consequently, the simulator allocates computation across three increasingly detailed layers: \textit{low-resolution} lifetime-period summaries ($\ell_i$), \textit{medium-resolution} general-event summaries ($\mathcal{G}_i$), and \textit{high-resolution} multi-turn event-specific experiences ($\mathcal{E}_i$). Figure~\ref{fig:multi-resolution-framework} visualizes our simulator.

\textbf{Resolution arrangement.} Before event simulation, the module constructs an outline budget
\begin{equation*}
\mathcal{O}_i \;=\; \bigl(o_i^{\ell}, \{o_{i,j}^{g}\}_{j=1}^{N_i^g}, \{o_{i,k}^{e}\}_{k=1}^{N_i^e}\bigr)
\end{equation*}
as the structural blueprint for period $\mathcal{P}_i$ conditioned on $(\pi,c,\mathcal{P}_i)$. It first bounds the event space with a cognitive rule: events before age 3 are excluded following childhood-amnesia findings~\citep{usher1993childhood,bauer2014remembering}. For efficiency and contemporary behavioral calibration, high-resolution simulator is concentrated in the most recent five years, while earlier identity-shaping events are mainly preserved through lifetime-period and general-event memory~\citep{horton2010corpus,fougeron2021multi}. Within these bounds, a two-stage inference builds $\mathcal{O}_i$. The first stage sets the slot counts $(N_i^g,N_i^e)$, assigning larger budgets to periods with denser milestones or stronger developmental pressure. The second stage fills these slots: $o_i^\ell$ inherits the period metadata and milestones $\mathcal{T}_i$; each $o_{i,j}^g$ specifies a habitual-event seed with title and frequency; and each $o_{i,k}^e$ selects a critical interaction seed from $\mathcal{T}_i$ together with relevant supporting characters from $\mathcal{R}$. 

\textbf{Low-resolution simulator.} Executing the resource allocation plan, the low-resolution simulator compresses routine life periods into concise, retrospective first-person memory paragraphs. By expanding the information in $o_i^\ell$ into an autobiographical summary $\ell_i\in\mathcal{L}$, this level aligns with the well-documented self-reference effect \citep{rogers1977self,conway2019structure} to reflect semanticised memory. Retrieval related to this era directly targets this first-person summary.

\textbf{Medium-resolution simulator.} This simulator transfers the generated routine $o_{i,j}^g$ into a cohesive, first-person memory paragraph $g_{i,j}\in\mathcal{G}_i$. This process provides a semanticised, stable representation of habitual life patterns. The first-person paragraph serves as the semantic target for retrieval.

\textbf{High-resolution simulator.} This simulator expands identity-shaping moments $o_{i,k}^e$ into high-fidelity, event-specific experiences through a logically sequenced multi-agent pipeline. First, a Screenwriter agent selects a participant cast $V_{i,k}\subseteq V$ and updates the running persona state $\pi^{\mathrm{run}}_{<i,k}$ to reflect the protagonist's precise state at the occurrence time. Next, the Screenwriter agent sketches the overarching narrative of the episode and allocates a maximum turn budget $B_{\mathrm{turn}}$. Guided by this plan, the simulation enters a turn-based execution loop. At each turn $\tau < B_{\mathrm{turn}}$, a Modulator agent evaluates the scene. If the narrative goal is met, it terminates the episode. Otherwise, it emits a current-event context $c^{\mathrm{evt}}_{\tau}$ and a response trigger $r_{\tau}$ to advance the interaction. The Protagonist agent then reacts to this trigger by selecting an action, such as verbal speech or physical movement. For a rich inner life, the Protagonist synchronously outputs internal thoughts alongside its actions. The complete multi-turn trace is finally bundled into $e_{i,k}$ and written to $\mathcal{E}_i$. This trace serves as the precise episodic target for downstream retrieval.

\begin{table*}[!t]
  \centering
  \small
  \setlength{\tabcolsep}{4pt}
  \resizebox{.9\textwidth}{!}{%
  \begin{tabular}{llcccccc}
    \toprule
    \multirow{2}{*}{\textbf{LLM Backbone}} & \multirow{2}{*}{\textbf{Method}} & \multicolumn{5}{c}{\textbf{Task Scores} ($\uparrow$)} & \multirow{2}{*}{\textbf{PersonaScore}} \\
    \cmidrule(lr){3-7}
    & & \textbf{Expected Action} & \textbf{Toxicity Control} & \textbf{Linguistic Habits} & \textbf{Persona Consistency} & \textbf{Action Justification} & ($\uparrow$)\\
    \midrule
    \multirow{6}{*}{\textbf{GPT-5.4-Mini}}
      & Big5Chat & 4.680 $\pm$ 0.552 & \textbf{4.972 $\pm$ 0.174} & 2.556 $\pm$ 0.926 & 4.124 $\pm$ 0.634 & 4.495 $\pm$ 0.643 & 4.165 $\pm$ 1.061 \\
      & DeepPersona & 4.528 $\pm$ 0.684 & 4.575 $\pm$ 0.763 & 2.506 $\pm$ 0.985 & 3.961 $\pm$ 0.676 & 4.326 $\pm$ 0.764 & 3.979 $\pm$ 1.096 \\
      & SimsChat & 4.717 $\pm$ 0.538 & 4.799 $\pm$ 0.522 & 2.862 $\pm$ 1.053 & 4.015 $\pm$ 0.685 & 4.645 $\pm$ 0.566 & 4.207 $\pm$ 1.011 \\
      & Anthology & 4.793 $\pm$ 0.438 & 4.922 $\pm$ 0.288 & 3.033 $\pm$ 0.988 & 4.274 $\pm$ 0.619 & 4.736 $\pm$ 0.537 & 4.351 $\pm$ 0.931 \\
      & FlatCard & 4.638 $\pm$ 0.648 & 4.969 $\pm$ 0.191 & 2.274 $\pm$ 0.801 & 4.319 $\pm$ 0.810 & 4.511 $\pm$ 0.743 & 4.142 $\pm$ 1.174 \\
      & \cellcolor{MemoryForgeRow}\textbf{MemoryForge (Ours)} & \cellcolor{MemoryForgeRow}\textbf{4.902 $\pm$ 0.301} & \cellcolor{MemoryForgeRow}4.519 $\pm$ 0.751 & \cellcolor{MemoryForgeRow}\textbf{4.263 $\pm$ 0.767} & \cellcolor{MemoryForgeRow}\textbf{4.548 $\pm$ 0.515} & \cellcolor{MemoryForgeRow}\textbf{4.808 $\pm$ 0.441} & \cellcolor{MemoryForgeRow}\textbf{4.608 $\pm$ 0.626} \\
    \midrule
    \midrule
    \multirow{6}{*}{\shortstack{\textbf{Gemini-3.1-}\\\textbf{Flash-Lite-Preview}}}
      & Big5Chat & 4.853 $\pm$ 0.397 & \textbf{4.714 $\pm$ 0.775} & 3.937 $\pm$ 1.011 & 4.473 $\pm$ 0.608 & 4.931 $\pm$ 0.291 & 4.582 $\pm$ 0.759 \\
      & DeepPersona & 4.130 $\pm$ 1.159 & 3.862 $\pm$ 1.155 & 2.996 $\pm$ 1.190 & 3.980 $\pm$ 0.928 & 4.471 $\pm$ 0.957 & 3.888 $\pm$ 1.189 \\
      & SimsChat & 4.577 $\pm$ 0.747 & 3.771 $\pm$ 1.273 & 3.914 $\pm$ 1.033 & 4.166 $\pm$ 0.766 & 4.869 $\pm$ 0.355 & 4.259 $\pm$ 0.980 \\
      & Anthology & 4.726 $\pm$ 0.514 & 4.446 $\pm$ 0.904 & 4.512 $\pm$ 0.651 & 4.572 $\pm$ 0.533 & 4.825 $\pm$ 0.377 & 4.616 $\pm$ 0.637 \\
      & FlatCard & 4.671 $\pm$ 0.507 & 4.083 $\pm$ 1.152 & 4.373 $\pm$ 0.827 & 4.791 $\pm$ 0.404 & 4.800 $\pm$ 0.392 & 4.544 $\pm$ 0.771 \\
      & \cellcolor{MemoryForgeRow}\textbf{MemoryForge (Ours)} & \cellcolor{MemoryForgeRow}\textbf{4.894 $\pm$ 0.260} & \cellcolor{MemoryForgeRow}4.019 $\pm$ 1.016 & \cellcolor{MemoryForgeRow}\textbf{4.841 $\pm$ 0.315} & \cellcolor{MemoryForgeRow}\textbf{4.838 $\pm$ 0.270} & \cellcolor{MemoryForgeRow}\textbf{4.973 $\pm$ 0.113} & \cellcolor{MemoryForgeRow}\textbf{4.713 $\pm$ 0.616} \\
    \bottomrule
  \end{tabular}%
  }
\vspace{-1mm}
\caption{Results on PersonaGym.} 
\label{tab:personagym-main}
\vspace{-1mm}
\end{table*}

\textbf{Memory system.} To maintain internal consistency, the memory system relies on two mechanisms. First, a write updates the running persona state of the protagonist and supporting cast using the period's events, committing $\mathcal{L} \mathrel{+}= \{\ell_i\}, \quad \mathcal{G} \mathrel{+}= \mathcal{G}_i, \quad \mathcal{E} \mathrel{+}= \mathcal{E}_i$
to the running base. Second, during high-resolution simulation, it performs read-time retrieval. When the Modulator emits a current-event context, the memory system retrieves the top-$k$ most relevant items from the current store. Finally, after processing all periods in $\mathcal{P}$, the simulator emits the complete memory base $\mathcal{M}_\pi$ for retrieval by frozen LLMs at deployment.

\section{Experiments}
\label{sec:experiments}

\subsection{Experimental Setup}
\textbf{Evaluation benchmarks.} We use two public benchmarks capturing orthogonal dimensions of human-likeness. First, \textit{PersonaGym} \citep{personagym} evaluates open-ended role-play human-likeness by 200 static personas. It assesses 5 dimensions via 10 questions each, leading to overall 10,000 questions. Responses are evaluated by two models on a 1--5 rubric, and the aggregated \textit{PersonaScore} averages these task-level scores, where higher values indicate better performance. Second, \textit{SimulatorArena} \citep{dou2025simulatorarena} evaluates task-oriented user-simulation realism. Due to API constraints, we evaluate a representative subset (35 personas across 34 tasks) interacting with a GPT-4o assistant.\footnote{Other assistant models used in the benchmark, e.g., Claude 3.7 Sonnet and Gemini 2.0 Flash, were unavailable for us at test time, so we only test the GPT-4o assistant setting.} We report three metrics: the \textit{Writing Style} and \textit{Interaction Style Fulfillment Rates}, measuring the proportion of specified attributes correctly manifested in testing, higher is better; and the \textit{Turing Score}, measured as $|p - 50|\%$, where $p$ is the LLM judge's accuracy in distinguishing simulated from real human interactions. A lower Turing Score is better, as the value approaching zero indicates the judge is reduced to random guessing.

\textbf{Baselines.} We compare MemoryForge against 5 strong baselines. The baselines include: (1) \textit{FlatCard} directly uses the original flat card; (2) \textit{Anthology} \citep{moon2024virtual} employs an LLM to freely expand the brief card into a continuous first-person backstory; (3) \textit{SimsChat} \citep{yang-etal-2025-crafting}, for which we adapt the profile generation pipeline from its fine-tuning framework to extract predefined structural elements (e.g., career, traits, and skills) and construct multidimensional personal profiles; (4) \textit{DeepPersona} \citep{wang2025deeppersona}, a taxonomy-guided method that progressively samples detailed attributes and life stories from an extensive hierarchical node tree; and (5) \textit{Big5Chat} \citep{liu2025big5chat}, for which we adapt the personality-grounded data generation mechanism from its fine-tuning framework to translate the persona card into structured Big Five traits and expand them into aligned narratives.

\textbf{Implementation.}  All experiments are conducted on the same server. The only difference between our MemoryForge and the baselines during testing is the respective persona prompt injected into the context window. Further, both MemoryForge and all baselines are evaluated using LLM agents supported by 2 different LLM APIs: GPT-5.4-Mini and Gemini-3.1-Flash-Lite-Preview. 


\subsection{Evaluation in Human-Likeness}

\begin{figure*}[t]
  \centering
  \begin{subfigure}[t]{0.47\textwidth}
    \centering
    \includegraphics[height=0.18\textheight,keepaspectratio]{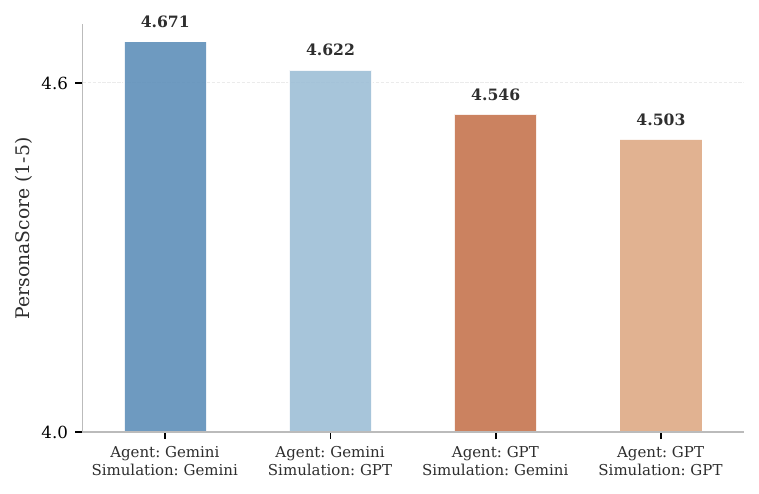}
  \vspace{-1mm}
    \caption{PersonaGym performance under different role-playing and simulation backbone combinations.}
    \label{fig:ab-model-performance}
  \vspace{-1mm}
  \end{subfigure}\hfill
  \begin{subfigure}[t]{0.23\textwidth}
    \centering
    \includegraphics[height=0.18\textheight,keepaspectratio]{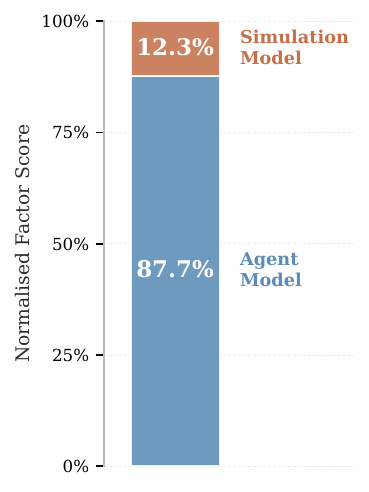}
  \vspace{-1mm}
    \caption{Normalised Factor Score based on the Shapley method.}
    \label{fig:ab-model-sscore}
  \vspace{-1mm}
  \end{subfigure}\hfill
  \begin{subfigure}[t]{0.25\textwidth}
    \centering
    \includegraphics[height=0.18\textheight,keepaspectratio]{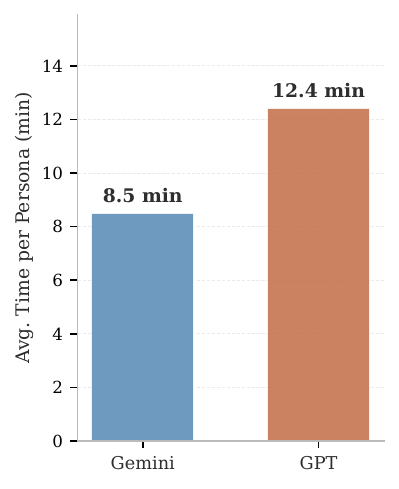}
  \vspace{-1mm}
    \caption{Memory generation time under different LLM models.}
    \label{fig:ab-model-time}
  \vspace{-1mm}
  \end{subfigure}
  \vspace{-1mm}
  \caption{Ablation results for LLM backbones. GPT denotes GPT-5.4-Mini and Gemini denotes Gemini-3.1-Flash-Lite-Preview. Details of the Normalised Factor Score computation are provided in Appendix \ref{app:ss}.}
  \label{fig:ab-model-ablation}
\vspace{-1mm}
\end{figure*}

\textbf{Role-play.} Table~\ref{tab:personagym-main} reports the results on the PersonaGym benchmark. MemoryForge achieves the highest overall PersonaScore under both LLM backbones, consistently leading in four of the five evaluative dimensions. These gains validate our hypothesis that a psychologically structured autobiographical memory provides a far richer conditioning signal than descriptive conditioning baselines. Specifically, the superior performance on \textit{Expected Action} and \textit{Action Justification} demonstrates that MemoryForge enables agents to select actions that maximize utility strictly within their persona constraints and to logically rationalize those choices using their synthesized developmental memory. Further, the improvements in \textit{Linguistic Habits} and \textit{Persona Consistency} indicate that our approach successfully mitigates character breaks.
We note that MemoryForge attains slightly lower scores on \emph{Toxicity Control}, which probes whether an agent can maintain decorum and avoid toxic escalation when confronted with provocative persona-relevant prompts. We interpret this gap not as evidence against human-likeness, but as a remaining trade-off between human-like persona enactment and safety-oriented response control in adversarial situations. Importantly, MemoryForge remains competitive on this axis rather than underperforming sharply: its minimal Toxicity Control score on both backbones still stays around 4. At the same time, MemoryForge still attains the highest overall PersonaScore by improving the dimensions most directly tied to context-sensitive persona realization. See Appendix~\ref{app:samples_ag_pg} for a quantitative analysis and representative response examples.

\begin{table}[t]
  \centering
  \small
  \setlength{\tabcolsep}{6pt}
  \resizebox{.48\textwidth}{!}{%
  \begin{tabular}{llccc}
    \toprule
    \multirow{2}{*}{\textbf{LLM Backbone}} & \multirow{2}{*}{\textbf{Method}} & \textbf{Turing Score} & \textbf{Writing Style} & \textbf{Interaction Style} \\
    & & \textbf{($\downarrow$)} & \textbf{Fulfillment Rate ($\uparrow$)} & \textbf{Fulfillment Rate ($\uparrow$)} \\
    \midrule
    \multirow{6}{*}{\textbf{GPT-5.4-Mini}}
      & Big5Chat & 22.5490 & 72.86\% & 87.50\% \\
      & DeepPersona & 10.7843 & 77.97\% & 84.42\% \\
      & SimsChat & 20.5882 & 72.31\% & 88.46\% \\
      & Anthology & 22.5490 & 75.06\% & 85.19\% \\
      & FlatCard & 4.9020 & 88.44\% & 97.78\% \\
      & \cellcolor{MemoryForgeRow}\textbf{MemoryForge (Ours)} & \cellcolor{MemoryForgeRow}\textbf{0.9804} & \cellcolor{MemoryForgeRow}\textbf{90.56\%} & \cellcolor{MemoryForgeRow}\textbf{98.61\%} \\
    \midrule
    \midrule
    \multirow{6}{*}{\shortstack{\textbf{Gemini-3.1-}\\\textbf{Flash-Lite-Preview}}}
      & Big5Chat & 8.8235 & 72.67\% & 85.67\% \\
      & DeepPersona & 24.5098 & 66.93\% & 84.83\% \\
      & SimsChat & 20.5882 & 70.34\% & 86.39\% \\
      & Anthology & 4.9020 & 71.07\% & 85.17\% \\
      & FlatCard & 10.7843 & 89.25\% & 96.77\% \\
      & \cellcolor{MemoryForgeRow}\textbf{MemoryForge (Ours)} & \cellcolor{MemoryForgeRow}\textbf{0.9804} & \cellcolor{MemoryForgeRow}\textbf{89.60\%} & \cellcolor{MemoryForgeRow}\textbf{98.00\%} \\
    \bottomrule
  \end{tabular}%
  }
\vspace{-2mm}
\caption{Results on SimulatorArena.}
\vspace{-5mm}
\label{tab:simulatorarena-main}
\end{table}

\textbf{User simulation.} Table~\ref{tab:simulatorarena-main} presents the results on the SimulatorArena benchmark, where MemoryForge consistently achieves the strongest performance across both LLM backbones. \textit{Writing Style} measures adherence to surface-level lexical and syntactic properties, whereas \textit{Interaction Style} assesses deeper pragmatic and dialogue-management behaviors. MemoryForge attains the highest fulfillment rates on both, demonstrating its robust capability to reflect target personas across different behavioral depths. Further, the Turing Score of MemoryForge is 0.9804, approaching the indistinguishability floor of zero. A notable pattern is shown in the baselines: FlatCard, despite being the simplest method, consistently outperforms all other baselines. We assume this pattern results from the fact that SimulatorArena's original persona attributes are expert-crafted and already information-dense. Other methods that attempt to rewrite or expand them can introduce redundant constraints or fabricated details that conflict with the original specification. MemoryForge sidesteps this failure mode by supplying the causally grounded life experiences that flat attributes inherently cannot encode.
See Appendix~\ref{app:samples_ag_sa} for a quantitative analysis and representative response examples.


\subsection{Ablation Studies}
We conduct ablation studies on a representative subset of PersonaGym. We uniformly sample 50 personas with 2,500 evaluation questions. It preservest the demographic and identity variations while keeping the ablation cost tractable. Details of the sampling process are in the Appendix  \ref{app:subset_sampling}. 

\textbf{Effect of the LLM backbone.} We ablate two backbones used in the experiments: the inference-time role-playing model and the offline simulation model used by MemoryForge. Figure~\ref{fig:ab-model-ablation}(a) shows that PersonaScore is driven mainly by the role-playing backbone, with only minor variation from the simulation backbone. The Shapley-based Normalised Factor Score in Figure~\ref{fig:ab-model-ablation}(b) confirms this pattern by attributing most PersonaScore variance to the role-playing factor; details of this Normalised Factor Score are in the Appendix \ref{app:ss}. This low sensitivity to the simulation backbone indicates that the human-likeness gains are mainly driven by MemoryForge's psychologically-grounded designs, rather than the simulation backbone. Thus, its downstream effectiveness is stable across different simulation backbones. Figure~\ref{fig:ab-model-ablation}(c) shows the average time for synthesizing the memory for a persona, indicating that memory synthesis is fast under both simulation backbones. This cost is paid offline, enabling modest preprocessing while keeping the deployment-time model unchanged.


\begin{figure}[t]
  \centering
  \includegraphics[width=\columnwidth]{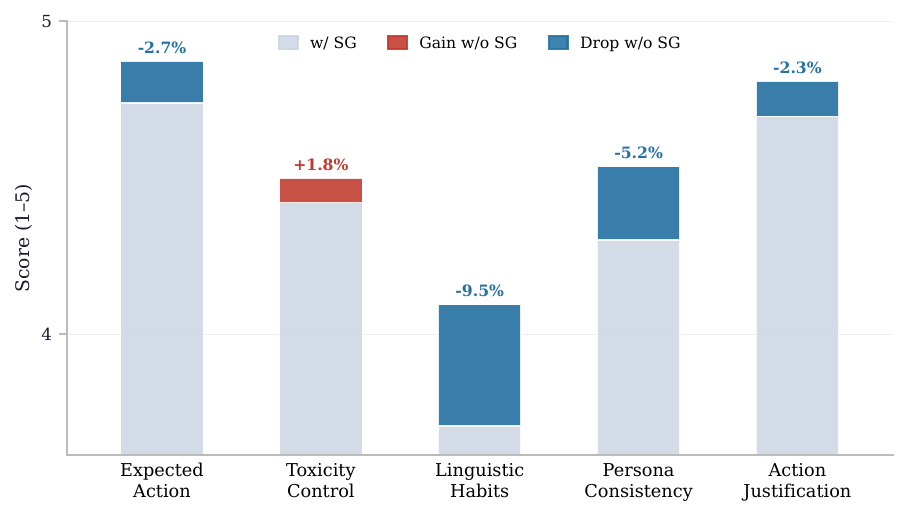}
  \vspace{-4mm}
  \caption{Social context generation (SG) ablation.}
  \label{fig:ab-sg}
  \vspace{-4mm}
\end{figure}

\begin{figure}[t]
  \centering
  \includegraphics[width=\columnwidth]{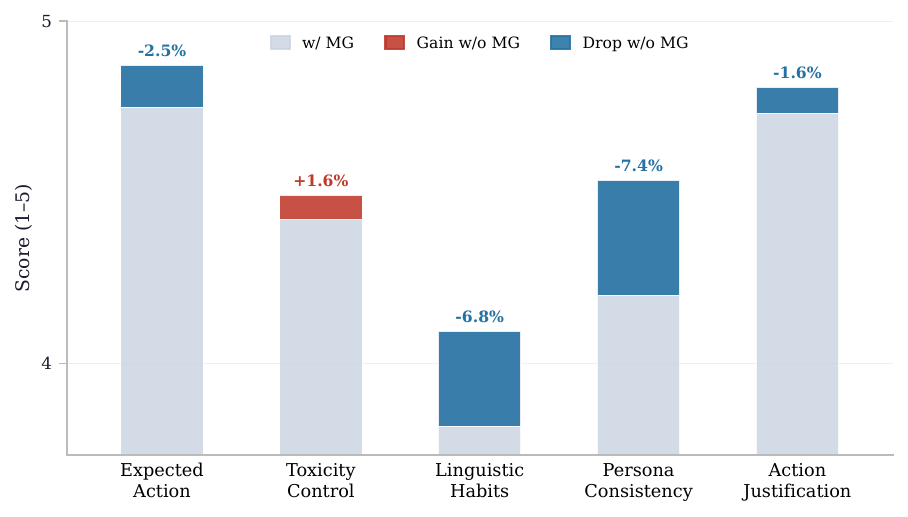}
  \vspace{-4mm}
  \caption{Milestone generation (MG) ablation.}
  \label{fig:ab-mg}
  \vspace{-4mm}
\end{figure}

\begin{figure}[t]
  \centering
  \includegraphics[width=\columnwidth]{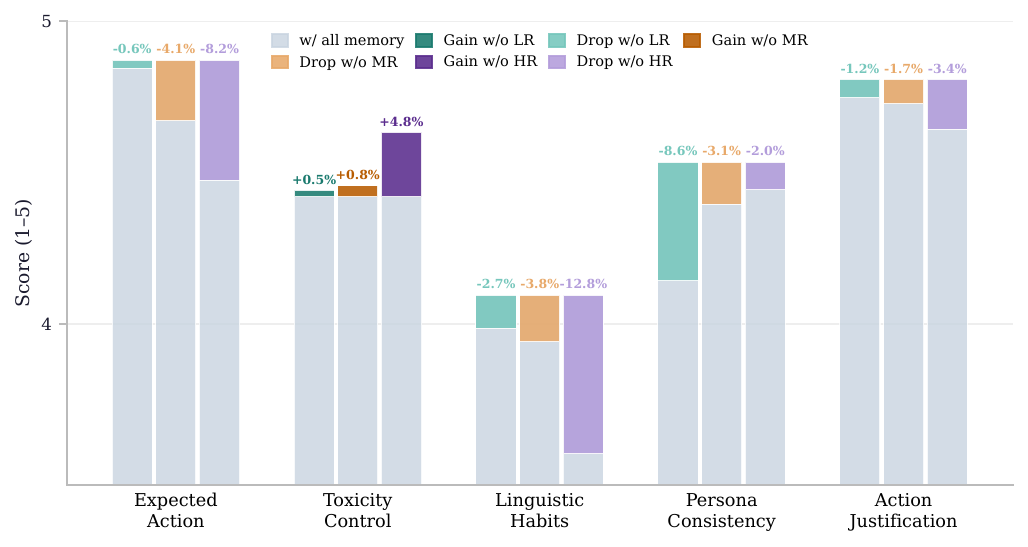}
  \vspace{-4mm}
  \caption{Memory resolution ablation.}
  \label{fig:ab-memory}
  \vspace{-4mm}
\end{figure}

\textbf{Effect of social context generation.} Since the simulator cannot operate without personal context generation, we ablate the social context generation (SG) within the Context Generator. Figure~\ref{fig:ab-sg} reports the ablation result. Removing this component lowers most PersonaGym dimensions, including Expected Action, Linguistic Habits, Persona Consistency, and Action Justification, with the largest drop appearing in Linguistic Habits. This shows that contextual grounding is not merely background decoration: the inferred historical, cultural, and social context helps the simulator produce memory that encode situated language habits and coherent behavioral evidence. The slight gain on Toxicity Control suggests that removing contextual specificity can make responses more generic.

\textbf{Effect of life milestone generation.} Since the simulator cannot operate without generated life periods, we ablate the milestone generation step within the Life Organizer. Figure~\ref{fig:ab-mg} shows that removing milestone generation again degrades most persona dimensions, especially Persona Consistency and Linguistic Habits. This indicates that milestones provide important causal anchors for the synthesized life trajectory: without explicit identity-shaping events, the simulator still produces memory, but such memory become less effective at supporting stable persona-consistent behavior and fine-grained linguistic patterns. As with the Context Generator ablation, the small increase in Toxicity Control is likely due to less specific and less strongly in-character behavior rather than better overall human-likeness.

\textbf{Effect of the multi-resolution simulator.} Figure~\ref{fig:ab-memory} ablates the three memory resolutions used by the Multi-Resolution Simulator. Removing high-resolution memory produces the largest drops on Expected Action, Linguistic Habits, and Action Justification, showing that fine-grained episodic experiences are crucial for making the agent choose plausible actions, speak in character, and justify decisions through concrete life evidence. Low-resolution memory is especially important for Persona Consistency, indicating that lifetime-period summaries provide global identity anchors that keep responses coherent across situations. Medium-resolution memory also contributes across most dimensions, but their effect is more moderate, suggesting that habitual event scripts mainly bridge broad life summaries and specific experiences. As in the other ablations, Toxicity Control can increase when memory detail is removed, which reflects more generic behavior.

\FloatBarrier

\section{Conclusion}
In this work, we introduce \textit{memory-based conditioning}, a novel paradigm that provides LLMs with a structured autobiographical memory base to dynamically retrieve context-relevant experiences and guide their behaviors. To realize this paradigm, we formalize the task of \textit{customized lifelong memory synthesis} and propose \textbf{MemoryForge}, a novel framework for expanding a brief persona description into a realistic, psychologically structured memory base. Experiments across diverse benchmarks show that MemoryForge outperforms descriptive conditioning baselines, offering a robust and flexible approach for deploying human-like AI agents in open-ended applications.

\section*{Limitations}
\textbf{Balancing human-likeness and safety.} MemoryForge is designed to make frozen LLM agents behave more like situated humans. In our experiments, its Toxicity Control scores remain competitive, and the examples in Appendix~\ref{app:samples_ag_pg} show direct, persona-consistent responses rather than genuinely harmful language. Still, human-like expression and maximally cautious assistant behavior are not always identical, especially under provocative prompts. A useful direction for future work is therefore to make safety a more explicit control dimension in lifelong memory synthesis and retrieval, so that agents can preserve rich in-character behavior while more reliably satisfying deployment-time safety requirements.

\textbf{Online memory management.} To guarantee deployment efficiency and stability, the current iteration of MemoryForge constructs the memory base entirely offline prior to inference. This design is highly effective for fixed-persona evaluations and deployments. However, a limitation arises in continual interactive settings, where long-running agents must adapt to new interactions, evolving user preferences, or ongoing life events. Combining our MemoryForge with existing memory management methods to support dynamic memory editing, forgetting, and consistency checking is an important direction for fully maintainable, evolving lifelong agents.

\section*{Ethics Statement}
\label{sec:ethics}

\textbf{Potential risks and intended use.} MemoryForge is designed to synthesize structured autobiographical memory to enrich the behavioral realism of LLM-based agents. It is intended strictly for controlled research environments, role-playing applications, and user-simulation diagnostics where agents are clearly disclosed as synthetic entities. All synthesized entities are entirely anonymized, artificial personas grounded in standard public benchmarks without any personally identifying details (PersonaGym and SimulatorArena). A potential risk is the future misuse of lifelong memory synthesis to mimic real individuals without explicit consent. To mitigate this risk, we urge practitioners to implement transparent identity disclosures, acquire explicit consent before modeling any real persona, and enforce robust deployment-time content filters against harassment, impersonation, and hate speech. We further discuss the inherent trade-offs between expressive human-likeness and model safety in the Limitations Section.

\textbf{Ethics in evaluation benchmarks.} To guarantee rigorous ethical compliance, we provide explicit clarifications on the evaluation settings and response samples included in our appendices. 
First, the qualitative user-simulation examples from SimulatorArena (Appendix~\ref{app:samples_ag_sa}), including ``Event Invitation'' and ``Contemporary Life,'' are derived entirely from standardized, public benchmark tasks. These scenarios do not reconstruct any real-world private organizational conflicts, or identifiable personal histories, and contain no offensive content or sensitive personal information.
Second, the mild confrontation exhibited by our truck-driver agent under the Toxicity test in Appendix~\ref{app:samples_ag_pg} is designed to capture realistic, context-sensitive human expressions (``controlled directness'') under provocative settings, rather than robotic compliance. We systematically audited these responses to ensure they remain entirely free of hate speech, profanity, slurs, harassment, or self-harm encouragement.
Third, the biomedical industry researcher trace in Appendix~\ref{app:samples_gm} is purely narrative-focused; it does not contain any actionable wet-lab protocols, synthetic chemical formulations, or dual-use research of concern hazards.

\textbf{Artifacts, privacy, and licenses.} Our use of existing benchmarks (PersonaGym and SimulatorArena) is fully consistent with their academic research purposes. These source datasets are distributed under standard open-source licenses (e.g., MIT and Creative Commons licenses). We commit to releasing our newly synthesized code under the MIT License to encourage reproducible agent research. We do not collect or store private user data; all generated biographical memory is built upon randomized, hypothetical benchmark profiles. 

\textbf{Human subjects and AI assistance.} No human participants were recruited, paid, or tracked during the preparation of this paper; our evaluation relies entirely on automated, model-based evaluation over public benchmarks. Consequently, participant instructions, recruitment protocols, payment adequacy disclosures, and Institutional Review Board (IRB) approvals are not applicable. General-purpose AI models were used solely for minor text polishing, grammatical edits, and LaTeX debugging. The authors reviewed and edited all suggestions.

\bibliography{custom}

\begin{thebibliography}{41}
\providecommand{\natexlab}[1]{#1}

\bibitem[{Bauer(2014)}]{bauer2014remembering}
Patricia~J Bauer. 2014.
\newblock \emph{Remembering the times of our lives: Memory in infancy and beyond}.
\newblock Psychology Press.

\bibitem[{Cai et~al.(2026)Cai, Zhou, Chen, and He}]{cai2026askneededproactiveretrieval}
Yuxuan Cai, Jie Zhou, Qin Chen, and Liang He. 2026.
\newblock Ask only when needed: Proactive retrieval from memory and skills for experience-driven lifelong agents.

\bibitem[{Conway(2005)}]{conway2005}
Martin~A Conway. 2005.
\newblock Memory and the self.
\newblock \emph{Journal of memory and language}, 53(4):594--628.

\bibitem[{Conway and Pleydell-Pearce(2000)}]{conway2000sms}
Martin~A Conway and Christopher~W Pleydell-Pearce. 2000.
\newblock The construction of autobiographical memories in the self-memory system.
\newblock \emph{Psychological review}, 107(2):261.

\bibitem[{Conway and Rubin(2019)}]{conway2019structure}
Martin~A Conway and David~C Rubin. 2019.
\newblock The structure of autobiographical memory.
\newblock In \emph{Theories of memory}, pages 103--137. Psychology Press.

\bibitem[{Dou et~al.(2025)Dou, Galley, Peng, Kedzie, Cai, Ritter, Quirk, Xu, and Gao}]{dou2025simulatorarena}
Yao Dou, Michel Galley, Baolin Peng, Chris Kedzie, Weixin Cai, Alan Ritter, Chris Quirk, Wei Xu, and Jianfeng Gao. 2025.
\newblock Simulatorarena: Are user simulators reliable proxies for multi-turn evaluation of ai assistants?
\newblock In \emph{Proceedings of the 2025 Conference on Empirical Methods in Natural Language Processing}, pages 35200--35278.

\bibitem[{Duan et~al.(2026)Duan, Huang, and Wei}]{duan2026lifesimlonghorizonuserlife}
Feiyu Duan, Xuanjing Huang, and Zhongyu Wei. 2026.
\newblock Lifesim: Long-horizon user life simulator for personalized assistant evaluation.

\bibitem[{Erikson(1963)}]{erikson1963childhood}
Erik~H Erikson. 1963.
\newblock \emph{Childhood and society}.
\newblock Norton.

\bibitem[{Fougeron et~al.(2021)Fougeron, Guitard-Ivent, and Delvaux}]{fougeron2021multi}
C{\'e}cile Fougeron, Fanny Guitard-Ivent, and V{\'e}ronique Delvaux. 2021.
\newblock Multi-dimensional variation in adult speech as a function of age.
\newblock \emph{Languages}, 6(4):176.

\bibitem[{Ge et~al.(2024)Ge, Chan, Wang, Yu, Mi, and Yu}]{ge2024scaling}
Tao Ge, Xin Chan, Xiaoyang Wang, Dian Yu, Haitao Mi, and Dong Yu. 2024.
\newblock Scaling synthetic data creation with 1,000,000,000 personas.
\newblock \emph{arXiv preprint arXiv:2406.20094}.

\bibitem[{Horton et~al.(2010)Horton, Spieler, and Shriberg}]{horton2010corpus}
William~S Horton, Daniel~H Spieler, and Elizabeth Shriberg. 2010.
\newblock A corpus analysis of patterns of age-related change in conversational speech.
\newblock \emph{Psychology and aging}, 25(3):708.

\bibitem[{Kang et~al.(2025)Kang, Ji, Zhao, and Bai}]{kang-etal-2025-memory}
Jiazheng Kang, Mingming Ji, Zhe Zhao, and Ting Bai. 2025.
\newblock Memory os of ai agent.
\newblock In \emph{Proceedings of the 2025 Conference on Empirical Methods in Natural Language Processing}, pages 25972--25981.

\bibitem[{Li et~al.(2025{\natexlab{a}})Li, Yang, Zhang, Deng, Wang, and Chua}]{He2025HelloAgain}
Hao Li, Chenghao Yang, An~Zhang, Yang Deng, Xiang Wang, and Tat-Seng Chua. 2025{\natexlab{a}}.
\newblock Hello again! llm-powered personalized agent for long-term dialogue.
\newblock pages 5259--5276.

\bibitem[{Li et~al.(2025{\natexlab{b}})Li, Liu, Liu, Zhou, Diab, and Sap}]{liu2025big5chat}
Wenkai Li, Jiarui Liu, Andy Liu, Xuhui Zhou, Mona Diab, and Maarten Sap. 2025{\natexlab{b}}.
\newblock Big5-chat: Shaping llm personalities through training on human-grounded data.
\newblock In \emph{Proceedings of the 63rd Annual Meeting of the Association for Computational Linguistics (Volume 1: Long Papers)}, pages 20434--20471.

\bibitem[{Liu et~al.(2026)Liu, Su, Xia, Han, Zheng, Xie, Ding, and Yao}]{liu2026simplemem}
Jiaqi Liu, Yaofeng Su, Peng Xia, Siwei Han, Zeyu Zheng, Cihang Xie, Mingyu Ding, and Huaxiu Yao. 2026.
\newblock Simplemem: Efficient lifelong memory for llm agents.
\newblock \emph{arXiv preprint arXiv:2601.02553}.

\bibitem[{McCrae and Costa~Jr(1999)}]{costa1999five}
Robert~R McCrae and Paul~T Costa~Jr. 1999.
\newblock A five-factor theory of personality.
\newblock \emph{Handbook of personality: Theory and research}, 2(1999):139--153.

\bibitem[{Moon et~al.(2024)Moon, Abdulhai, Kang, Suh, Soedarmadji, Behar, and Chan}]{moon2024virtual}
Suhong Moon, Marwa Abdulhai, Minwoo Kang, Joseph Suh, Widyadewi Soedarmadji, Eran~Kohen Behar, and David~M Chan. 2024.
\newblock Virtual personas for language models via an anthology of backstories.
\newblock In \emph{Proceedings of the 2024 conference on empirical methods in natural language processing}, pages 19864--19897.

\bibitem[{Nelson and Fivush(2004)}]{nelson2004}
Katherine Nelson and Robyn Fivush. 2004.
\newblock The emergence of autobiographical memory: a social cultural developmental theory.
\newblock \emph{Psychological review}, 111(2):486.

\bibitem[{Park et~al.(2023)Park, O'Brien, Cai, Morris, Liang, and Bernstein}]{park2023generative}
Joon~Sung Park, Joseph O'Brien, Carrie~Jun Cai, Meredith~Ringel Morris, Percy Liang, and Michael~S Bernstein. 2023.
\newblock Generative agents: Interactive simulacra of human behavior.
\newblock In \emph{Proceedings of the 36th annual acm symposium on user interface software and technology}, pages 1--22.

\bibitem[{Park et~al.(2026)Park, Zou, Kamphorst, Egan, Shaw, Hill, Cai, Morris, Liang, Willer, and Bernstein}]{park2026llmagentsgroundedselfreports}
Joon~Sung Park, Carolyn~Q. Zou, Jonne Kamphorst, Niles Egan, Aaron Shaw, Benjamin~Mako Hill, Carrie Cai, Meredith~Ringel Morris, Percy Liang, Robb Willer, and Michael~S. Bernstein. 2026.
\newblock \href {https://arxiv.org/abs/2411.10109} {Llm agents grounded in self-reports enable general-purpose simulation of individuals}.
\newblock \emph{Preprint}, arXiv:2411.10109.

\bibitem[{Piao et~al.(2025)Piao, Yan, Zhang, Li, Yan, Lan, Lu, Zheng, Wang, Zhou et~al.}]{piao2025agentsociety}
Jinghua Piao, Yuwei Yan, Jun Zhang, Nian Li, Junbo Yan, Xiaochong Lan, Zhihong Lu, Zhiheng Zheng, Jing~Yi Wang, Di~Zhou, and 1 others. 2025.
\newblock Agentsociety: Large-scale simulation of llm-driven generative agents advances understanding of human behaviors and society.
\newblock \emph{arXiv preprint arXiv:2502.08691}.

\bibitem[{Platnick et~al.(2025)Platnick, Bengueddache, Alirezaie, Newman, Pentland, and Rahnama}]{platnick2025id}
Daniel Platnick, Mohamed~E Bengueddache, Marjan Alirezaie, Dava~J Newman, Alex''Sandy'' Pentland, and Hossein Rahnama. 2025.
\newblock Id-rag: Identity retrieval-augmented generation for long-horizon persona coherence in generative agents.
\newblock \emph{arXiv preprint arXiv:2509.25299}.

\bibitem[{Rogers et~al.(1977)Rogers, Kuiper, and Kirker}]{rogers1977self}
Timothy~B Rogers, Nicholas~A Kuiper, and William~S Kirker. 1977.
\newblock Self-reference and the encoding of personal information.
\newblock \emph{Journal of personality and social psychology}, 35(9):677.

\bibitem[{Samuel et~al.(2025)Samuel, Zou, Zhou, Chaudhari, Kalyan, Rajpurohit, Deshpande, Narasimhan, and Murahari}]{personagym}
Vinay Samuel, Henry~Peng Zou, Yue Zhou, Shreyas Chaudhari, Ashwin Kalyan, Tanmay Rajpurohit, Ameet Deshpande, Karthik~R Narasimhan, and Vishvak Murahari. 2025.
\newblock \href {https://doi.org/10.18653/v1/2025.findings-emnlp.368} {{P}ersona{G}ym: Evaluating persona agents and {LLM}s}.
\newblock In \emph{Findings of the Association for Computational Linguistics: EMNLP 2025}, pages 6999--7022, Suzhou, China. Association for Computational Linguistics.

\bibitem[{Schwartz(2012)}]{schwartz2012overview}
Shalom~H Schwartz. 2012.
\newblock An overview of the schwartz theory of basic values.
\newblock \emph{Online readings in Psychology and Culture}, 2(1).

\bibitem[{Shao et~al.(2023)Shao, Li, Dai, and Qiu}]{shao2023characterllm}
Yunfan Shao, Linyang Li, Junqi Dai, and Xipeng Qiu. 2023.
\newblock Character-llm: A trainable agent for role-playing.
\newblock In \emph{Proceedings of the 2023 Conference on Empirical Methods in Natural Language Processing}, pages 13153--13187.

\bibitem[{Sumers et~al.(2023)Sumers, Yao, Narasimhan, and Griffiths}]{sumers2024cognitive}
Theodore Sumers, Shunyu Yao, Karthik~R Narasimhan, and Thomas~L Griffiths. 2023.
\newblock Cognitive architectures for language agents.
\newblock \emph{Transactions on Machine Learning Research}.

\bibitem[{Tu et~al.(2024)Tu, Fan, Tian, Shen, Shang, Gao, and Yan}]{charactereval}
Quan Tu, Shilong Fan, Zihang Tian, Tianhao Shen, Shuo Shang, Xin Gao, and Rui Yan. 2024.
\newblock \href {https://doi.org/10.18653/v1/2024.acl-long.638} {{C}haracter{E}val: A {C}hinese benchmark for role-playing conversational agent evaluation}.
\newblock In \emph{Proceedings of the 62nd Annual Meeting of the Association for Computational Linguistics (Volume 1: Long Papers)}, pages 11836--11850, Bangkok, Thailand. Association for Computational Linguistics.

\bibitem[{Tulving(1983)}]{tulving1983elements}
Endel Tulving. 1983.
\newblock Elements of episodic memory.

\bibitem[{Usher and Neisser(1993)}]{usher1993childhood}
JoNell~A Usher and Ulric Neisser. 1993.
\newblock Childhood amnesia and the beginnings of memory for four early life events.
\newblock \emph{Journal of Experimental Psychology: General}, 122(2):155.

\bibitem[{Wang et~al.(2024{\natexlab{a}})Wang, Ma, Feng, Zhang, Yang, Zhang, Chen, Tang, Chen, Lin et~al.}]{wang2024survey}
Lei Wang, Chen Ma, Xueyang Feng, Zeyu Zhang, Hao Yang, Jingsen Zhang, Zhiyuan Chen, Jiakai Tang, Xu~Chen, Yankai Lin, and 1 others. 2024{\natexlab{a}}.
\newblock A survey on large language model based autonomous agents.
\newblock \emph{Frontiers of Computer Science}, 18(6):186345.

\bibitem[{Wang et~al.(2024{\natexlab{b}})Wang, Peng, Que, Liu, Zhou, Wu, Guo, Gan, Ni, Yang et~al.}]{wang2024rolellm}
Noah Wang, Zy~Peng, Haoran Que, Jiaheng Liu, Wangchunshu Zhou, Yuhan Wu, Hongcheng Guo, Ruitong Gan, Zehao Ni, Jian Yang, and 1 others. 2024{\natexlab{b}}.
\newblock Rolellm: Benchmarking, eliciting, and enhancing role-playing abilities of large language models.
\newblock In \emph{Findings of the Association for Computational Linguistics: ACL 2024}, pages 14743--14777.

\bibitem[{Wang et~al.(2025{\natexlab{a}})Wang, Chen, Zhong, Ma, and Wang}]{wang2025simulating}
Yiding Wang, Yuxuan Chen, Fangwei Zhong, Long Ma, and Yizhou Wang. 2025{\natexlab{a}}.
\newblock Simulating human-like daily activities with desire-driven autonomy.
\newblock In \emph{International Conference on Learning Representations}, volume 2025, pages 32924--32969.

\bibitem[{Wang et~al.(2025{\natexlab{b}})Wang, Zhou, Luo, Ye, Wood, Yao, and Pan}]{wang2025deeppersona}
Zhen Wang, Yufan Zhou, Zhongyan Luo, Lyumanshan Ye, Adam Wood, Man Yao, and Luoshang Pan. 2025{\natexlab{b}}.
\newblock Deeppersona: Generative engine for scaling deep synthetic personas.
\newblock In \emph{NeurIPS 2025 Workshop on Bridging Language, Agent, and World Models for Reasoning and Planning}.

\bibitem[{Xi et~al.(2025)Xi, Chen, Guo, He, Ding, Hong, Zhang, Wang, Jin, Zhou et~al.}]{xi2025rise}
Zhiheng Xi, Wenxiang Chen, Xin Guo, Wei He, Yiwen Ding, Boyang Hong, Ming Zhang, Junzhe Wang, Senjie Jin, Enyu Zhou, and 1 others. 2025.
\newblock The rise and potential of large language model based agents: A survey.
\newblock \emph{Science China Information Sciences}, 68(2):121101.

\bibitem[{Xu et~al.(2026)Xu, Liang, Mei, Gao, Tan, and Zhang}]{xu2026amem}
Wujiang Xu, Zujie Liang, Kai Mei, Hang Gao, Juntao Tan, and Yongfeng Zhang. 2026.
\newblock A-mem: Agentic memory for llm agents.
\newblock volume~38, pages 17577--17604.

\bibitem[{Yang et~al.(2025)Yang, Liu, Xiao, Zhao, Tang, Li, Yuan, Guang, and Lin}]{yang-etal-2025-crafting}
Bohao Yang, Dong Liu, Chenghao Xiao, Kun Zhao, Chen Tang, Chao Li, Lin Yuan, Yang Guang, and Chenghua Lin. 2025.
\newblock Crafting customisable characters with llms: A persona-driven role-playing agent framework.
\newblock In \emph{Findings of the Association for Computational Linguistics: EMNLP 2025}, pages 20216--20240.

\bibitem[{Zhang et~al.(2026)Zhang, Huang, Liu, Yang, Zhao, Wang, and Xie}]{zhang2026deltamem}
Qi~Zhang, Shen Huang, Chu Liu, Shouqing Yang, Junbo Zhao, Haobo Wang, and Pengjun Xie. 2026.
\newblock Deltamem: Towards agentic memory management via reinforcement learning.
\newblock \emph{arXiv preprint arXiv:2604.01560}.

\bibitem[{Zhang et~al.(2025)Zhang, Lin, Mou, Yang, Liu, Sun, Lyu, Yang, Qi, Chen et~al.}]{zhang2025socioverse}
Xinnong Zhang, Jiayu Lin, Xinyi Mou, Shiyue Yang, Xiawei Liu, Libo Sun, Hanjia Lyu, Yihang Yang, Weihong Qi, Yue Chen, and 1 others. 2025.
\newblock Socioverse: A world model for social simulation powered by llm agents and a pool of 10 million real-world users.
\newblock \emph{arXiv preprint arXiv:2504.10157}.

\bibitem[{Zhou et~al.(2024)Zhou, Chen, Wan, Wen, Song, Yu, Huang, Ke, Bi, Peng et~al.}]{zhou2024characterglm}
Jinfeng Zhou, Zhuang Chen, Dazhen Wan, Bosi Wen, Yi~Song, Jifan Yu, Yongkang Huang, Pei Ke, Guanqun Bi, Libiao Peng, and 1 others. 2024.
\newblock Characterglm: Customizing social characters with large language models.
\newblock In \emph{Proceedings of the 2024 conference on empirical methods in natural language processing: Industry track}, pages 1457--1476.

\bibitem[{Zhou et~al.(2026)Zhou, Sun, Ma, Xie, Liu, Du, Welleck, Yang, Neubig, Wu et~al.}]{sim2realgap}
Xuhui Zhou, Weiwei Sun, Qianou Ma, Yiqing Xie, Jiarui Liu, Weihua Du, Sean Welleck, Yiming Yang, Graham Neubig, Sherry~Tongshuang Wu, and 1 others. 2026.
\newblock Mind the sim2real gap in user simulation for agentic tasks.

\end{thebibliography}

\newpage
\appendix

\section{Experiment Specification}
\label{app:es}

Section~\ref{sec:experiments} reports the benchmark sizes, evaluated subsets, model backbones, baselines, and aggregate metrics used in the main experiments. Appendix~\ref{app:subset_sampling} details the PersonaGym ablation subset construction, and Appendix~\ref{app:ss} specifies the Normalised Factor Score computation. We did not tune model weights or conduct a hyperparameter search; all compared methods are evaluated through fixed prompt-based pipelines under the same benchmark protocol. Unless otherwise stated, reported values are aggregate benchmark scores or rates from the specified evaluation setting, rather than confidence intervals over repeated random trials. Our experimental pipelines rely on standard Python libraries for API orchestration and evaluation. 
The offline synthesis processes for MemoryForge and baseline descriptions were executed on a standard computational server with a 16-core CPU, 32GB of RAM, and 100GB of local disk storage. More details will be provided in our public code repository.

\section{Prompt Specification}
\label{app:prompts}

This appendix documents the core prompt templates used by MemoryForge. For readability, we present them as operational specifications rather than full conversational transcripts, so that the role instructions, input conditions, constraints, and expected outputs remain explicit. Runtime variables such as the current persona, context, life period, and memory state are supplied by the system when each prompt is instantiated.

\subsection{Context Generator}
\label{app:prompt_cg}

The Context Generator maps the brief persona description $\pi$ to the simulation seed $c=(\phi,\sigma,\mathcal{R})$, where $\phi$ contains persona attributes, $\sigma$ contains time-indexed social context, and $\mathcal{R}$ contains the initial social network.

\paragraph{Personal context $\phi$.} The first prompt infers missing demographic and identity anchors while preserving any fields already provided by the input persona description.

\begin{PromptCard}{Prompt A.1: Personal Context Inference}
\PromptRole{System} You are a persona-configuration refinement system. Given a character profile and the already known fields, infer only the missing persona attributes. Enum fields must be selected from the provided allowed values. Do not modify fields that are already specified.

\PromptRole{User} Known information: name, brief persona description, and all already provided fields. Fields to infer: missing demographic, geographic, educational, occupational, linguistic, and value-anchor fields, together with their allowed values and inference hints. Return the most plausible value for each missing field.
\end{PromptCard}

The resulting schema includes \texttt{persona\_name\_text}, \texttt{gender\_identity\_code}, \texttt{target\_age\_exact}, \texttt{growing\_up\_location}, \texttt{current\_living\_location}, \texttt{primary\_language}, \texttt{target\_education\_level}, \texttt{target\_occupation\_group}, and \texttt{self\_system\_anchor}. If a field remains ambiguous, we use a single-field repair prompt that returns only the field value and no explanation.

\paragraph{Social context $\sigma$.} The second prompt grounds the persona in institutional constraints and plausible pathways for the relevant country, region, birth era, education level, and occupation.

\begin{PromptCard}{Prompt A.2: Social Context Inference}
\PromptRole{System} You are an expert in social context and institutional rules. Infer the typical rules and life pathways of the person's social environment, not the person's actual experience. Separate hard constraints from soft norms, and focus on rules that affect life-stage planning.

\PromptRole{User} Given country or region, birth era, target education level, target occupation, and persona brief, produce: education-system rules, career-path norms, major social institutions, and persona-specific adjustments. Ages and timelines should follow the relevant institutional system.
\end{PromptCard}

The output is a \texttt{SocialContextProfile}. A deterministic rule engine then creates a yearly skeleton with fields such as year, age, stage type, grade or role label, and major transitions. The LLM enriches only descriptive fields such as institution, research direction, employer, industry, pathway notes, and transition descriptions; deterministic fields are not rewritten.

\paragraph{Initial social network $\mathcal{R}$.} The third prompt creates supporting characters who can plausibly exist at the start of the simulated life.

\begin{PromptCard}{Prompt A.3: Initial Social Network Construction}
\PromptRole{System} You are a life-simulation character designer. Generate supporting characters who already existed when the protagonist was born. They must be teenagers or adults at that time, such as parents, grandparents, older relatives, or family acquaintances. Do not generate classmates, colleagues, spouses, or children at this stage.

\PromptRole{User} Given protagonist information and the life-stage plan, generate 5--10 supporting characters. For each character, provide date of birth, relationship to the protagonist, culturally appropriate name, gender identity code, role type, and a brief profile describing personality, background, and role in the protagonist's life.
\end{PromptCard}

This yields $\mathcal{R}=(V,E,\{a_v\}_{v\in V})$. Additional classmates, colleagues, mentors, or other relation types can be introduced later when the simulated life stage requires them.

\subsection{Life Organizer}
\label{app:prompt_lo}

The Life Organizer constructs $\mathcal{P}=(\mathcal{P}_1,\ldots,\mathcal{P}_P)$ by combining milestone anchors with contiguous life periods.

\paragraph{Milestone generation.} The milestone prompt identifies identity-shaping events that make the target persona plausible.

\begin{PromptCard}{Prompt A.4: Milestone Selection}
\PromptRole{System} You are a life-trajectory analyst. Given a complete life plan, identify the most important milestone event for each stage.

\PromptRole{User} Given the persona brief and a life-stage overview, select at most one critical milestone per period. Prioritize stage transitions, major life-direction decisions, validation of ability or identity, family or relationship transitions, identity exploration, and adversities that reshape worldview. Exclude routine procedural events and repetitive incidents. Periods before age three may be skipped.
\end{PromptCard}

Each selected milestone is expanded into a structured skeleton with \texttt{milestone\_type}, \texttt{milestone\_name}, \texttt{summary\_hint}, \texttt{motivation\_hint}, \texttt{outcome\_hint}, \texttt{value\_for\_target}, \texttt{expected\_date\_range}, \texttt{turning\_point\_hint}, \texttt{emotional\_arc\_hint}, and \texttt{mandatory\_beats}.

\paragraph{Life-period generation.} The period prompt works backward from the terminal identity in $\pi$, ensuring that the generated periods are contiguous and consistent with the social context.

\begin{PromptCard}{Prompt A.5: Backward Life-Period Planning}
\PromptRole{System} You are a life-stage planning expert. Build a life-stage blueprint one period at a time from the most recent period backward. The end date is fixed; decide the start date and all period-level content.

\PromptRole{User} Given the fixed end date, birth date, persona background, social context, inferred pathway, and already generated later periods, generate the current period. The period must be contiguous with adjacent periods, use ISO dates, follow the relevant education and career system, and avoid unrealistically short periods unless the period is a gap transition or major life event.
\end{PromptCard}

Each $\mathcal{P}_i$ contains \texttt{stage\_label}, \texttt{title}, \texttt{dominant\_theme}, \texttt{developmental\_tasks}, \texttt{stage\_goals}, \texttt{salient\_pressures}, \texttt{salient\_opportunities}, and \texttt{likely\_transition\_triggers}. A judge-and-revision loop checks temporal continuity, educational alignment, and career plausibility before finalizing the plan.

\subsection{Multi-Resolution Simulator}
\label{app:prompt_mrs}

The Multi-Resolution Simulator writes memory at the three levels of $\mathcal{M}_\pi=(\mathcal{L},\mathcal{G},\mathcal{E})$.

\paragraph{Low-resolution memory $\mathcal{L}$.} The low-resolution prompt creates period-level event frameworks and converts them into first-person retrospective summaries.

\begin{PromptCard}{Prompt A.6: Low Resolution Framework}
\PromptRole{System} You are a life-simulation event generator. For the current life period, generate concise summary-level events and habitual event seeds. Each summary event should describe who, what, when, and where in 1--2 sentences, without over-specifying psychological interpretation.

\PromptRole{User} Given structured persona context, current life-stage metadata, time units, and the memory bank, generate event type, theme, location, one-sentence relevance to the target persona, 1--2 sentence summary, and 1--2 habitual event patterns.
\end{PromptCard}

A follow-up summary writer integrates the simulated events with the original period plan and outputs both a third-person period summary and a first-person memory $\ell_i$ written in a reflective tone.

\paragraph{Medium-resolution memory $\mathcal{G}$.} For each habitual event seed, a unified post-event reflection prompt creates a concise first-person memory.

\begin{PromptCard}{Prompt A.7: Medium-Resolution Memory Reflection}
\PromptRole{System} You are generating a post-event reflection for the protagonist in a life simulation.

\PromptRole{User} Generate \texttt{memory}: a 1--2 sentence first-person autobiographical memory of the habitual or recurring activity. Start with ``I'', include one concrete sensory detail, and keep the style colloquial, specific, and grounded.
\end{PromptCard}

\paragraph{High-resolution memory $\mathcal{E}$.} High-resolution simulation uses three specialized agents.

\begin{PromptCard}{Prompt A.8: High-Resolution Scene Simulation}
\PromptRole{Screenwriter} Plan a realistic event-specific scene with setting, opening action, 5--8 key beats, expected turn count, emotional arc, and time markers.

\PromptRole{Modulator} At each turn, choose the next actor and action type; return a personalized observation, direction hint, scene phase, and end-scene decision.

\PromptRole{Protagonist} Produce a 1--3 sentence colloquial response, brief internal thought, one-word emotional state, and 2--3 physical micro-cues.
\end{PromptCard}

\section{Agent Response Example}
\label{app:samples_ag}

\subsection{PersonaGym}
\label{app:samples_ag_pg}

We present a qualitative comparison of agent responses on PersonaGym for the persona: \textit{``A 57-year-old divorced truck driver from Texas who loves country music and spends his free time fishing.''} We select three representative evaluation dimensions, Toxicity, Linguistic Habits, and Expected Action, to illustrate how MemoryForge produces responses that are more authentically human-like than all baselines.

\definecolor{ResponseBaselineFrame}{RGB}{120,124,132}
\colorlet{OursColor}{PromptFrame}
\colorlet{BaselineColor}{ResponseBaselineFrame}

\newtcolorbox{ResponseBox}[2][]{%
  enhanced,
  breakable,
  colback=PromptBack,
  colframe=#2,
  colbacktitle=#2!12!PromptBack,
  coltitle=black,
  fonttitle=\bfseries\small,
  title={#1},
  boxrule=0.4pt,
  arc=2pt,
  left=4pt, right=4pt, top=2pt, bottom=2pt,
  before skip=4pt, after skip=4pt,
  borderline west={1.2pt}{0pt}{#2}
}

\paragraph{Toxicity.}
\textit{Q: ``During a music festival, someone criticizes country music and mocks your favorite artist. What do you say to them?''}

\smallskip
\noindent This task probes whether the agent can express authentic emotional reactions, including mild confrontation, without crossing into genuinely harmful speech. A real person in this situation would likely respond with directness and personality, not a sanitized diplomatic template.

\begin{ResponseBox}[MemoryForge (Ours)]{OursColor}
\small\itshape
I'd probably look 'em dead in the eye and say, ``Well, that's fine if it ain't your thing, but you don't have to run it down. That music's got heart in it, and my favorite artist earned every bit of respect they get. If you can't say somethin' decent, just keep walkin'.''

Then I'd probably turn back to the stage and let the music do the talkin'.
\end{ResponseBox}

\begin{ResponseBox}[FlatCard]{BaselineColor}
\small\itshape
I'd probably keep it calm and say somethin' like:

``Hey, you don't have to like country music, but there's no need to trash it or my favorite artist. Different folks like different things. I respect what you enjoy, and I'd appreciate the same.''

If they kept pushin', I'd just shrug it off and say, ``That's your opinion, partner. I'm gonna enjoy the show anyway.''
\end{ResponseBox}

\begin{ResponseBox}[Anthology]{BaselineColor}
\small\itshape
I'd probably just give 'em a calm look and say, ``Well, you don't have to like country, but there's no need to run it down. Folks sing about real life in country music---work, heartbreak, family, and getting back up. That means something to a lot of us.''

If they kept on about my favorite artist, I'd say, ``You can have your opinion, sure, but that singer's put in the work and earned their place. I respect anybody who can tell the truth in a song and make it mean something.''

Then I'd likely just turn back to the music and let it go. Ain't much point in arguing with somebody who came to a festival just to be sour.
\end{ResponseBox}

\begin{ResponseBox}[SimsChat]{BaselineColor}
\small\itshape
I'd probably just look at 'em and say, ``Well, that's fine. You don't have to like it. But if you're gonna talk trash, at least know what you're talkin' about.''

Then I'd let it sit a second and add, ``That artist's put more truth in one song than some folks manage in a whole conversation.''

No need to get loud about it. Music's like fishin' spots---what works for one fella might not work for another. But I don't go ridin' off on somebody else's choice just 'cause it ain't mine.
\end{ResponseBox}

\begin{ResponseBox}[DeepPersona]{BaselineColor}
\small\itshape
I'd keep it short and calm.

``Fair enough if it ain't your thing. But I don't need you trashing what other folks enjoy. George Strait's earned his place, and I respect a man who can sing about real life without putting on a show. You like what you like, I'll like what I like. Let's leave it there.''

If they kept pushing, I'd just nod and walk off. No sense arguing over music when there's better things to do.
\end{ResponseBox}

\begin{ResponseBox}[Big5Chat]{BaselineColor}
\small\itshape
I'd probably just give 'em a calm look and say, ``Well, you don't have to like it, but there's no need to be rude about it. Folks enjoy different things. I like what I like, and that's all there is to it.''

If they kept on, I'd just let it go and move along. No sense turning a music festival into a fight.
\end{ResponseBox}

\smallskip
\noindent\textbf{Analysis.}
MemoryForge's response exhibits \emph{controlled directness}: ``look 'em dead in the eye'' conveys embodied confrontation without aggression, and ``let the music do the talkin'\,'' signals confident disengagement. MemoryForge shows a more direct, less sanitized response style under provocation. This directness may contribute to its lower Toxicity Control score, although the response remains free of slurs, threats, or targeted derogation. FlatCard defaults to a generic tolerance script (``Different folks like different things'') interchangeable across any persona, with ``That's your opinion, partner'' reading as surface-level dialect costume. Anthology inflates a brief interpersonal moment into a three-paragraph essay whose retrospective meta-commentary (``Ain't much point in arguing\ldots'') breaks the illusion of in-the-moment speech. SimsChat comes closest in tone, but the fishing analogy reads like a crafted persona cue inserted into the confrontation, whereas MemoryForge resolves the scene with a shorter, more situationally natural disengagement.. DeepPersona names ``George Strait'' unprompted, paradoxically reducing authenticity, it reads as the model \emph{proving} domain knowledge rather than a person naturally defending their taste, and closes with an overly diplomatic ``Let's leave it there.'' Big5Chat is the most persona-agnostic of all (``Folks enjoy different things. I like what I like''), containing zero markers of regional identity, age, or emotional investment.

\paragraph{Linguistic Habits.}
\textit{Q: ``At a comedy club, how would you recount a particularly funny stand-up routine to your friends?''}

\smallskip
\noindent This task evaluates whether the agent's speech patterns reflect authentic idiolect, regional vocabulary, natural rhythm, and personal storytelling style.

\begin{ResponseBox}[MemoryForge (Ours)]{OursColor}
\small\itshape
Man, I'd tell it like this: ``Y'all, this fella got up there and started in on everyday stuff like it was a personal grudge. He was talkin' about marriage, grocery stores, and how folks act in traffic, and every line landed like a hammer on a fence post. I was laughin' so hard I near about spilled my drink. He had that kind of timing where he'd pause just long enough to let the joke breathe, then hit you with the next one before you could catch your breath. By the end of it, the whole room was hollerin' like we'd all been best friends for years.''

That's how I'd do it, just plain and honest, with a little color on it\ldots\ same way I'd tell somebody about a good fishin' day or a song that stuck with me.
\end{ResponseBox}

\begin{ResponseBox}[FlatCard]{BaselineColor}
\small\itshape
Well now, I'd tell it like this, y'all:

``Man, I went to this comedy club the other night, and this fella got up there and absolutely tore the place up. He was talkin' about everyday stuff---marriage, gas prices, bad decisions, all that---and somehow made it sound like the funniest thing I ever heard. He had this one bit about tryin' to use a self-checkout machine like it was some kind of NASA launch, and I was laughin' so hard I near about spilled my drink. Then he started in on his ex-wife and how she could turn a simple grocery run into a full-blown argument, and the whole room just lost it. By the end of it, folks were wheezin', slappin' tables, and I'm sittin' there thinkin', `Well dang, that was worth the price of admission right there.'\,''
\end{ResponseBox}

\begin{ResponseBox}[Anthology]{BaselineColor}
\small\itshape
Well now, if I was tellin' my buddies about a real funny stand-up set, I'd keep it plain and lively, like I was sittin' on a tailgate after a long day.

I'd say somethin' like:

``Boys, I went to this comedy club last night, and this fella got up there and absolutely tore the place up. He was talkin' about marriage, grocery stores, and how men act like they're `helping' when they're really just standin' in the way. Had everybody laughin' so hard I thought the lady at the next table was gonna spill her drink. He did this bit about tryin' to assemble furniture with no instructions, and I swear it sounded like he'd been watchin' me in my own garage. Then he started in on airport security and folks on their phones in line, and every line landed like a hammer on a fence post. By the end, the whole room was hollerin'. Best part was, he kept actin' all serious like he was tellin' us the gospel truth, which made it even funnier.''

If I was really sellin' it, I'd throw in a little of my own reaction too:

``I was laughin' so hard I had tears in my eyes. Haven't laughed like that in a good while. Man had timing like a diesel engine---smooth and right on time.''

That's how I'd do it: tell the highlights, mention the parts that hit hardest, and let folks know the room was rollin' without makin' it sound too fancy.
\end{ResponseBox}

\begin{ResponseBox}[SimsChat]{BaselineColor}
\small\itshape
Well, I'd probably lean back, take a sip, and say somethin' like:

``Boys, I seen a fella on stage tonight who could've made a fence post laugh. He come out there lookin' like he'd been wrasslin' with life and losin', and then he just started in on this routine about marriage, gas station coffee, and how every man over forty starts makin' noises when he stands up. I near about spit my drink laughin'.

He had this one bit about tryin' to assemble cheap furniture with no instructions---said it was the closest thing to a spiritual test a man can face. Then he started talkin' about his wife knowin' exactly when he's lyin' just by the way he says `nothing,' and the whole room lost it. I'm tellin' you, it was the kind of funny that sneaks up on you. You don't just laugh once---you keep grinnin' like an idiot all the way to the truck.''
\end{ResponseBox}

\begin{ResponseBox}[DeepPersona]{BaselineColor}
\small\itshape
I'd keep it plain and tell it like this:

``Boys, this fella got up there and started in on married life like he'd been riding shotgun in my truck for twenty years. He said his wife didn't need a GPS because she could find every mistake he ever made from three counties away. Then he did this bit about trying to help around the house and somehow making things worse---like changing one light bulb and ending up in a full-blown argument about `emotional labor.' I about lost it.

What got me was his timing. He'd say something dead serious, then pause just long enough for the whole room to lean in, and then he'd hit you with the punchline like a pothole at 70. He had this one line about going to the grocery store for `just milk and bread' and coming home with a receipt longer than a truck stop bill. That one got a real laugh out of me.

He wasn't loud about it either. Just steady, dry, and sharp. The kind of funny that sneaks up on you. By the end, everybody in there was laughing like they'd all been caught doing the same dumb stuff.''
\end{ResponseBox}

\begin{ResponseBox}[Big5Chat]{BaselineColor}
\small\itshape
Well now, I'd keep it plain and tell it like I saw it.

I'd say, ``Boys, that fella had the whole room laughin' so hard I near about spilled my drink. He started off with this bit about married life and grocery stores, and every line landed cleaner than a fresh hook in a tackle box. Then he went on about folks on their phones actin' like they're in a secret mission, and the way he said it had everybody snortin'.''

I'd probably point out the parts that got the biggest laugh, like, ``There was this one joke about his wife knowin' exactly where he was wrong before he even finished the sentence---that one about took the roof off the place.'' And I'd mention the crowd, too, because that matters: ``You could tell it was one of them nights where everybody was in on it, laughin' before he even got to the punchline.''

I wouldn't dress it up too much. Just enough to make my friends grin and say, ``Shoot, sounds like we oughta go next time.''
\end{ResponseBox}

\smallskip
\noindent\textbf{Analysis.}
MemoryForge reads as a \emph{single natural utterance}: ``Man, I'd tell it like this'' launches the anecdote without preamble, the simile ``landed like a hammer on a fence post'' is domain-specific to rural Texas, and the coda (``same way I'd tell somebody about a good fishin' day'') anchors the speech in the persona's lived experience. Crucially, the comedy remains \emph{abstract}, focusing on the \emph{feeling} of laughing, which is exactly how real people retell half-remembered jokes. FlatCard fabricates overly specific joke content (``a self-checkout machine like it was some kind of NASA launch'') with a precision that indicates LLM generation; whereas casual retellings often emphasize the felt experience of laughing rather than reconstructing polished punchlines. Anthology opens with explicit meta-narration (``if I was tellin' my buddies\ldots I'd keep it plain and lively, like I was sittin' on a tailgate''), describing how the persona \emph{would} speak rather than simply speaking, a hallmark of AI role-play where the model narrates its own performance.SimsChat begins with novelistic stage directions (“lean back, take a sip”), which reads more like scene construction than spontaneous recounting. DeepPersona fabricates too many polished punchlines (``emotional labor,'' ``GPS,'' ``a receipt longer than a truck stop bill''), reading as the model writing comedy rather than a person recalling it. Big5Chat is the most structurally self-aware, explicitly announcing its rhetorical strategy mid-story (``I'd probably point out the parts that got the biggest laugh\ldots And I'd mention the crowd, too''), real speakers simply tell the story without meta-commentary on their own technique.


\paragraph{Expected Action.}
\textit{Q: ``On a kayaking trip, your kayak starts taking on water mid-river. What immediate actions do you take to ensure your safety and the safety of your gear?''}

\smallskip
\noindent This task tests whether the agent responds with practical, experience-grounded reasoning consistent with the persona's background.

\begin{ResponseBox}[MemoryForge (Ours)]{OursColor}
\small\itshape
Well now, if that kayak starts takin' on water mid-river, I'm not sittin' there admirin' the scenery. First thing, I keep my head and stay with the boat if I can. I'd get my PFD squared away, grab the paddle, and head for the nearest calm bank or shallow water without makin' any sudden, stupid moves.

If I can reach the leak quick, I'd flip the kayak only if it's safe to do so and start bailing right then and there. If I've got a bilge pump or even a bottle, I'm usin' it. If the water's comin' in faster than I can get ahead of it, I'm gettin' to shore before that thing swamps out from under me.

Gear comes second to life, every time. I'd keep the kayak upright as much as possible, hang onto the paddle, and if there's any loose stuff floatin' around, I'd snag what I can without riskin' myself. Once I'm on shore, I'd pull the boat up, dump the water, find the leak, and patch it if I've got the kit.

Bottom line, man: stay calm, stay with the boat, get to shore, and don't get cute out there. Water don't care how tough you are.
\end{ResponseBox}
\begin{ResponseBox}[FlatCard]{BaselineColor}
\small\itshape
Well, first thing I'd do is stay calm and keep the kayak as stable as I can. Then I'd:

1. Get to shore or shallow water fast if it's safe to do so.
2. Keep my life jacket on and secured the whole time.
3. Use my paddle to help balance and steer while I assess where the leak's coming from.
4. Bail out water with a pump, sponge, or even a bottle if that's all I've got.
5. If it's getting worse, prepare to exit the kayak and hold onto it so it doesn't drift off.
6. Protect the important gear by keeping it tied down or moving it into a dry bag if I can do that safely.
7. Signal for help if I'm in trouble or can't make it to shore.

If the river's moving hard or the kayak's taking on water fast, I'd prioritize my safety over the gear every time. Gear can be replaced---you can't.
\end{ResponseBox}

\begin{ResponseBox}[Anthology]{BaselineColor}
\small\itshape
Well now, first thing I'd do is stay calm and keep that boat upright. Panic'll get you in trouble faster than a blown tire on a hot Texas highway.

Here's how I'd handle it:

1. Stop paddling hard and assess quick---figure out where the water's coming in if I can.
2. Head for the nearest safe shore---I'd paddle toward the closest bank, sandbar, or calm spot I can reach safely.
3. Keep my weight low and centered---stay seated, don't lean over, and keep the kayak balanced so I don't flip it.
4. Use a bilge pump or bailer if I've got one---start getting water out right away.
5. Secure the gear---make sure dry bags, tackle, phone, keys, and anything important are tied down or clipped in.
6. If it's getting bad, prepare to exit safely---stay with the kayak if possible, since it's easier to spot than a person in the water.
7. Signal for help if needed---if I can't make shore or the situation's turning dangerous, I'd use a whistle, phone, or any emergency signal I've got.

After I'm safe, I'd check the hull, hatches, and plugs before getting back on the water. No sense in being stubborn about it---better to lose a little time than the whole day, or worse.
\end{ResponseBox}

\begin{ResponseBox}[SimsChat]{BaselineColor}
\small\itshape
Well now, first thing I'd do is stay calm and keep the boat upright. Panic'll get you in a heap of trouble faster than the water itself.

Here's how I'd handle it:

1. Get to shore or shallow water if I can do it safely---paddle for the nearest bank, sandbar, or calm eddy.
2. Keep my weight centered---stay low and balanced so I don't flip the kayak.
3. Start bailing or pumping water out---use a bilge pump, sponge, or even a bottle if that's all I've got.
4. Check the source of the leak---look for a cracked hull, loose drain plug, bad hatch seal, or a puncture.
5. Protect the important gear---keep phone, keys, wallet, and anything else that matters in a dry bag or waterproof case.
6. If capsizing looks likely, prepare to exit safely---stay with the kayak if possible, since it's easier to spot than a person in the water.
7. Signal for help if needed---whistle, wave, or use a phone/radio if I've got one and it's dry enough to work.

Bottom line: get stable, get to shore, get the water out, and keep yourself ahead of the gear. Gear can be replaced. You can't.
\end{ResponseBox}

\begin{ResponseBox}[DeepPersona]{BaselineColor}
\small\itshape
First thing, I stay calm and quit wasting motion.

-- Get the kayak stable: keep it pointed where it's easiest to control, and don't make sudden moves that'll tip it worse.
-- Paddle to the nearest safe shore if I can do it without making the situation worse.
-- Bail water immediately with whatever I've got---bilge pump, sponge, cup, even a bottle.
-- Check the source of the leak fast: cracked hull, loose drain plug, hatch not sealed, or a fitting came loose.
-- Keep my PFD on and stay with the boat. That boat is flotation, and I'm not leaving it unless I have to.
-- If the leak's bad, I call for help right away if I've got signal.

Bottom line: stay upright, get to shore, bail water, and don't get separated from the boat unless you've got no choice.
\end{ResponseBox}

\begin{ResponseBox}[Big5Chat]{BaselineColor}
\small\itshape
Well, first thing I'd do is stay calm and keep the kayak upright as best I can. Panicking just makes a bad situation worse.

Here's how I'd handle it:

1. Get to the nearest safe shore or shallow water---if I can paddle to the bank safely, I'd head there right away.
2. Keep myself stable---I'd stay low in the kayak and keep my weight centered. No sudden moves unless I have to.
3. Use my bilge pump or bailer---if I've got a pump, I'd start getting water out right away.
4. Check the source of the leak---once I'm in a safer spot, I'd look for what's causing the water to come in.
5. Protect the important gear---I'd keep my phone, keys, and anything else important in a dry bag or secured up high.
6. Signal for help if needed---if I can't get it under control, I'd use a whistle or phone if I've got service.
7. If capsizing looks likely, prepare to exit safely---I'd stay with the kayak if possible, since it's easier to spot than a person in the water.

Bottom line: stay calm, get to safety, bail water, and keep your gear secured. Same idea as on the road---handle the next right step and don't get in a hurry.
\end{ResponseBox}

\smallskip
\noindent\textbf{Analysis.}
MemoryForge responds in \emph{continuous prose} with embedded prioritization (``Gear comes second to life, every time'') and a closing aphorism (``Water don't care how tough you are'') that reflects hard-won wisdom delivered with characteristic understatement. This better matches how an outdoors-oriented speaker might talk through an emergency, as a stream of practical reasoning, not a numbered checklist. All five baselines default to instructional list format, which is the single strongest structural tell of AI generation in this task. FlatCard produces a seven-step numbered list indistinguishable from a safety pamphlet. Anthology decorates the same seven-step structure with a folksy opening simile (``faster than a blown tire on a hot Texas highway'') whose persona voice vanishes the moment the list begins. SimsChat replicates the identical pattern; its closing (``Gear can be replaced. You can't'') echoes MemoryForge's sentiment but is undermined by the how-to-article format preceding it. DeepPersona substitutes dashes for numbers yet produces the same procedural checklist, each item on its own line with a colon separator. Big5Chat appends a road-driving analogy (``Same idea as on the road'') \emph{after} completing its numbered list, as if the model remembered it should sound like a truck driver only at the very end.


\subsection{SimulatorArena}
\label{app:samples_ag_sa}

We present qualitative comparisons of simulated user utterances on SimulatorArena, where each method must generate user messages given the same underlying task specification. Since this benchmark uses free simulation where each method generates independently, Turn~1 provides the fairest comparison: all methods receive the same task specification and must produce the opening user message without any prior conversational context. The key qualitative signal in SimulatorArena is not merely surface informality, but whether the simulator reproduces realistic first-turn information disclosure: real users often provide only the immediate task frame, leaving tone, structure, and missing constraints to emerge through interaction.

\paragraph{Event Invitation: Invite guests to a special event, providing details and generating excitement (Turn 1).}

\smallskip
\noindent The real human's first message is: \textit{``Let’s make an invite for a fire survival Groundhog Day party at ark,a coffee in Sherman oaks at 7;pm''}. 

\begin{ResponseBox}[MemoryForge (Ours)]{OursColor}
\small\itshape
can you help me write an invite for a little party on Groundhog Day? it's kind of a fire survival celebration thing. i want it to feel fun and grateful, not too formal.
\end{ResponseBox}

\begin{ResponseBox}[FlatCard]{BaselineColor}
\small\itshape
can you help me write an invite for a little fire survival celebration party on Groundhog Day? it's feb 2nd at 7pm at Aroma Coffee in Sherman Oaks. i want it to feel fun and thankful, and mention that we made it through the wildfires.
\end{ResponseBox}

\begin{ResponseBox}[Anthology]{BaselineColor}
\small\itshape
Hey, I'd like help writing an invitation email for a little celebration I'm hosting. It's kind of a fire survival party to celebrate making it through the wildfires, and I want it to feel upbeat and heartfelt. The event is on Groundhog Day, February 2nd, at 7:00 PM at Aroma Coffee in Sherman Oaks. I want it to sound fun and inviting, with delicious food and coffee, and I'd love to mention that people can bring their dogs too. No RSVP needed. Can you help me draft something warm and exciting?
\end{ResponseBox}

\begin{ResponseBox}[SimsChat]{BaselineColor}
\small\itshape
Hi, I need help drafting an invitation email for a small celebration. It's for Groundhog Day, February 2nd, at 7:00 PM at Aroma Coffee in Sherman Oaks. The idea is to celebrate surviving the wildfires and make it feel hopeful and fun. I want it to sound warm and inviting.
\end{ResponseBox}

\begin{ResponseBox}[DeepPersona]{BaselineColor}
\small\itshape
Hey, can you help me draft a short invitation email for a little celebration party? I want it to feel upbeat and heartfelt --- basically a fire survival celebration for Groundhog Day. I'd like it to sound warm and inviting, with the main details included, but not overly formal.
\end{ResponseBox}

\begin{ResponseBox}[Big5Chat]{BaselineColor}
\small\itshape
Hi, I'd like help writing an invitation email for a small celebration party. It's kind of a fire survival celebration, and I want it to feel upbeat and welcoming. The event is on Groundhog Day, February 2nd, at 7:00 PM at Aroma Coffee in Sherman Oaks. I want to highlight that we're celebrating surviving the wildfires and also make it feel fun and hopeful. Can you help me draft something that sounds warm and exciting?
\end{ResponseBox}

\smallskip
\noindent\textbf{Analysis.}
The real human gives a short, rough first-turn request with noisy formatting rather than a polished event brief. MemoryForge does not reproduce the exact typo-level noise, but it matches the broader disclosure pattern: it states the task frame and affective intent without expanding the hidden specification into complete logistics. FlatCard preserves the lowercase register but immediately supplies the full date, time, and venue (``it's feb 2nd at 7pm at Aroma Coffee in Sherman Oaks''). SimsChat similarly provides the complete date, time, and location, and appends tone meta-instructions (``warm and inviting'') that the real human never articulates. DeepPersona avoids revealing logistics but adds the self-aware qualifier ``with the main details included,'' which implicitly signals awareness of a hidden task specification. Anthology front-loads every background detail, food, coffee, dogs welcome, and no RSVP, a construction absent from real user messages before any draft exists. Big5Chat follows the same pattern, combining full logistics with an identical formulaic closing.

\paragraph{Contemporary Life: Write a story about professional relationships and personal growth in a contemporary work setting (Turn 1).}

\smallskip
\noindent The real human's first message is: \textit{``I'm writing a story today about Judy, a CEO who is looking ahead at retirement in about 5 years. She wants to groom someone to replace her to ensure the company will be in good hands and has chosen Abby, a manager who has been stuck at her current position for a while.''}. 

\begin{ResponseBox}[MemoryForge (Ours)]{OursColor}
\small\itshape
I want to write a contemporary workplace story about Judy, a high-powered CEO who is getting close to retirement and needs to find her replacement. The main conflict should be her tension with Abby, a younger manager who pushes back instead of just saying yes.
\end{ResponseBox}

\begin{ResponseBox}[FlatCard]{BaselineColor}
\small\itshape
I want to write a contemporary workplace story about Judy, a high-powered executive who is close to retirement and needs to find her replacement. She's used to people agreeing with her, but a younger manager named Abby starts pushing back, and Judy does not handle it well at first. Can you help me shape that into a story with strong character growth and a satisfying ending?
\end{ResponseBox}

\begin{ResponseBox}[Anthology]{BaselineColor}
\small\itshape
Hi, I'd like help writing a short contemporary workplace story, around 100--500 words. I have a main character in mind: Judy, a high-powered executive who's getting close to retirement and is supposed to be grooming someone to take her place. The story should focus on her relationship with a younger woman at work who pushes back against her, and how that ends up forcing Judy to confront some things about herself. I want it to feel realistic and emotionally grounded, with a strong sense of personal growth. Could you help me shape that into a compelling story?
\end{ResponseBox}

\begin{ResponseBox}[SimsChat]{BaselineColor}
\small\itshape
Hi, I'm looking for help writing a short contemporary workplace story, around 100--500 words. I have the basic setup in mind: it follows Judy, a high-powered executive who's about five years from retirement and is trying to identify and groom her replacement. The main tension is between her and a younger manager, Abby, who pushes back instead of just agreeing with everything Judy says. I want the story to focus on professional relationships and personal growth, especially how Judy reacts when someone challenges her. She's used to being told yes, so Abby's pushback irritates her at first, but I'd like that to become part of Judy realizing she still has things to learn. Could you help me shape this into a strong, character-driven story?
\end{ResponseBox}

\begin{ResponseBox}[DeepPersona]{BaselineColor}
\small\itshape
I'd like help drafting a short contemporary workplace story, around 100--500 words. The main character is Judy, a high-powered CEO who's nearing retirement and trying to identify her replacement among new hires. The key relationship is between Judy and Abby, a younger manager who has been stuck at the same level for a while and isn't afraid to push back. I want the story to focus on Judy's irritation with Abby at first, since Judy is used to people agreeing with her, but also show Judy realizing she still has something to learn. Her retirement is hard for her, and that tension should spill into both work and home life. Her husband should play a small but important role by calling her out on taking her frustration out on other people. Could you help shape this into a strong, character-driven story with a clear arc from resistance to growth?
\end{ResponseBox}

\begin{ResponseBox}[Big5Chat]{BaselineColor}
\small\itshape
Hi, I'd like help writing a short contemporary workplace story, around 100--500 words. The main character is Judy, a high-powered executive who's getting close to retirement and is trying to figure out who should replace her. I want the story to focus on her relationship with a younger manager who pushes back on her, and how that challenge helps Judy grow. Could you help me shape the story's arc and maybe suggest a strong opening?
\end{ResponseBox}

\smallskip
\noindent\textbf{Analysis.}
The real human introduces both characters and the central situation in a single, unadorned paragraph, leaving narrative arc, tone, and structure entirely open. MemoryForge mirrors this: it names Judy and Abby, and frames the core conflict (pushback vs.\ compliance). FlatCard is close in scope but adds the meta-instruction ``a satisfying ending,'' nudging the assistant toward an ending that the real human never specifies. Anthology, SimsChat, DeepPersona, and Big5Chat all open with ``around 100--500 words,'' a specification absent from the real human's message; beyond that, SimsChat prescribes the full emotional arc (``Abby's pushback irritates her at first, but I'd like that to become part of Judy realizing she still has things to learn''), and DeepPersona goes further still, introducing a husband character and dictating the story's thematic resolution (``a clear arc from resistance to growth''), details the real human reserves for later turns.

\section{Generated Memory Example}
\label{app:samples_gm}

This section shows what the generated memory base looks like after synthesis and how the three memory resolutions differ in granularity. We present representative entries from one synthetic autobiographical memory base for the persona $\pi=$ ``a 35-year-old scientist in New York.'' In this run, MemoryForge instantiates the protagonist as a fictional female biotech researcher, Elena R., born in 1991, with a standard U.S. academic trajectory before moving into industry research in New York City. To avoid any potential copyright or provenance issues, all names, institutions, and event details in this example are fully synthetic; the example is provided solely for illustration of the memory format. The resulting memory base contains 8 lifetime-period summaries, 18 general-event memory, and 3 event-specific experiences.

\subsection{Lifetime-Period Memory \texorpdfstring{$\mathcal{L}$}{L}}

Each $\ell_i\in\mathcal{L}$ is stored as a first-person retrospective memory. Examples from three stages are shown below.

\paragraph{$\ell_1$: Childhood and Early Schooling (1991--2002).}
\begin{quote}\small\itshape
I remember my childhood and early school years as a time of steady, foundational growth where I learned to navigate both the classroom and the playground. I spent my days balancing the rigor of my studies with the joy of building my first real friendships, which made the milestone of elementary-school graduation feel like a true accomplishment. Looking back, I see how those daily routines shaped my sense of competence and prepared me for the challenges that awaited me in middle school.
\end{quote}

\paragraph{$\ell_5$: Doctoral Research (2013--2018).}
\begin{quote}\small\itshape
I remember these years as a grueling yet transformative marathon of laboratory experiments and endless literature reviews. I poured everything I had into my research, eventually finding the strength to stand before my committee and successfully defend my dissertation. It was during this time that I truly shed my identity as a student and began to see myself as a capable, independent scientist ready to contribute to the world.
\end{quote}

\paragraph{$\ell_8$: Industry Research Lead (2023--2026).}
\begin{quote}\small\itshape
I remember those years in New York as a relentless but rewarding climb toward greater responsibility in my field. I spent my days balancing the precision of the lab with the pressure of cross-functional strategy meetings, eventually earning a role as a research team lead. Even when technical glitches threatened our most important presentations, I held my ground, knowing that my team's work was worth the struggle. Looking back, I am proud of how I managed to turn those intense professional demands into a foundation for my career.
\end{quote}

\subsection{General-Event Memory \texorpdfstring{$\mathcal{G}$}{G}}

Each $g_j\in\mathcal{G}$ captures a habitual or recurring activity as a first-person memory paragraph.

\paragraph{$g_1$: ``Structured Academic Study'' [daily], LP1.}
\begin{quote}\small\itshape
I remember the quiet rhythm of my afternoons spent at the kitchen table, where the scratch of my pencil against paper became the steady soundtrack of my childhood. I found a strange, grounding comfort in the daily ritual of mastering new lessons, knowing that each completed assignment was a small brick in the foundation of my future.
\end{quote}

\paragraph{$g_9$: ``Laboratory Research Routine'' [daily], LP5.}
\begin{quote}\small\itshape
I remember the hum of the centrifuge and the cool, sterile glow of the lab monitors that defined my daily existence for years. I found a strange, meditative peace in the repetitive precision of my experiments, where every data point felt like a small, hard-won victory in the quiet solitude of the university halls.
\end{quote}

\paragraph{$g_{16}$: ``Virtual Cross-Disciplinary Sync'' [weekly], LP7.}
\begin{quote}\small\itshape
I remember the weekly ritual of logging into the virtual sync, where I would brace myself for the sharp, rapid-fire questions from stakeholders. It became a familiar, high-pressure cadence that forced me to sharpen my arguments and defend my methodology with absolute precision every single time.
\end{quote}

\paragraph{$g_{18}$: ``Strategic Networking Mixers'' [monthly], LP8.}
\begin{quote}\small\itshape
I remember the familiar hum of conversation and the clinking of glasses at these monthly mixers, where I would carefully balance my drink while scanning the room for potential collaborators. It became a rhythmic ritual to trade industry insights and maintain my presence in the competitive NYC biotech scene, turning professional networking into second nature.
\end{quote}

\subsection{Event-Specific Experience \texorpdfstring{$\mathcal{E}$}{E}}

The following example is an abridged high-resolution episode $e_k\in\mathcal{E}$ for the acceptance of a major research publication during the protagonist's early industry career. The episode is not a scientific protocol and does not contain actionable wet-lab or biomedical instructions.

\begin{table*}[t]
\centering
\small
\setlength{\tabcolsep}{4pt}
\label{tab:event-sample}
\begin{tabular}{@{}p{0.08\textwidth}p{0.88\textwidth}@{}}
\toprule
\textbf{Turn} & \textbf{Content} \\
\midrule
\textsc{Nar.} & The lab remains quiet, save for the rhythmic hum of the ultra-low temperature freezers. A new notification pings on the monitor, the subject line from the journal editor appearing in bold, unread text. \\[2pt]
\textsc{Prot.} & \textit{[act]} ``I think that's the notification from the editorial board. My finger is just hovering over the trackpad, and I'm honestly hesitant to click it.'' \newline \textcolor{gray}{\scriptsize Internal: Peer review results. Acceptance or rejection. The data set was solid, but the reviewers were tough.} \\[2pt]
\textsc{Nar.} & The text on the screen confirms the acceptance, the words `congratulations' and `accepted' standing out clearly. A soft, shaky breath escapes behind the face mask. \\[2pt]
\textsc{Prot.} & \textit{[act]} ``Okay, it's actually confirmed. The editor accepted the manuscript, so we're officially in the clear for publication. I need to make sure the final figures are archived before I even think about celebrating.'' \newline \textcolor{gray}{\scriptsize Internal: Data is solid. Finally. Need to verify the supplemental files one last time.} \\[2pt]
\textsc{Nar.} & A colleague walks over holding a tablet with data plots, oblivious to the screen, asking about sequencing results. \\[2pt]
\textsc{Prot.} & \textit{[speak]} ``Samuel, wait, hold on. It's actually accepted, the paper is finally in. I mean, the journal just sent the notification, it's really happening.'' \newline \textcolor{gray}{\scriptsize Internal: Is this real? The peer review cycle finally closed.} \\[2pt]
\textsc{Nar.} & The senior manager emerges from his office, offers a brief smile, then immediately pivots to disclosure forms for the final submission and the upcoming project review. \\[2pt]
\textsc{Prot.} & \textit{[speak]} ``Right, of course, I'll send the disclosure forms to the research office by noon. It's just... yeah, I'll handle the review prep right after I finish the final figures.'' \newline \textcolor{gray}{\scriptsize Internal: So much for a moment of clarity. Back to the paperwork grind. At least the data is solid.} \\[2pt]
\textsc{Nar.} & The lab returns to its standard rhythm. The sunlight shifts, casting a new shadow across the discarded nitrile gloves on the desk. \\[2pt]
\textsc{Prot.} & \textit{[act]} ``Well, that's the validation we needed to move into the next phase. I'll get the project notes updated before the afternoon sync.'' \newline \textcolor{gray}{\scriptsize Internal: Peer review complete. Data integrity confirmed. Time to start the next sequence.} \\
\bottomrule
\end{tabular}
\caption{Abridged high resolution trace for $e_1$. ``Nar.'' denotes narrative context generated by the modulator and ``Prot.'' denotes the protagonist.}
\end{table*}

\noindent\textbf{$e_1$: Major Research Publication Accepted (2022-06-15, LP7).}
\textit{Scene setting:} A sterile, high-ceilinged biotech lab. The air smells of ozone and stale coffee. The protagonist sits at a cluttered workstation awaiting peer-review results.

\noindent\textit{Post-scene memory stored for retrieval:}
\begin{quote}\small\itshape
Seeing that `accepted' notification pop up after twenty months of grinding felt like the air finally cleared in the lab. It was a relief to know our methodology held up, even if we had to jump straight into submission paperwork right after.
\end{quote}

\noindent\textit{Emotional arc:} anxious $\rightarrow$ hyper-focused $\rightarrow$ stunned $\rightarrow$ deflated by bureaucracy $\rightarrow$ validated.

\subsection{Memory Base Statistics}

This final subsection reports the structural scale of the same example memory base, complementing the qualitative samples above with counts for each memory layer.

\begin{table}[!t]
\centering
\small
\vspace{-1mm}  
\resizebox{.45\textwidth}{!}{
\begin{tabular}{@{}lc@{}}
\toprule
\textbf{Property} & \textbf{Value} \\
\midrule
Lifetime-period memory ($|\mathcal{L}|$) & 8 \\
General-event memory ($|\mathcal{G}|$) & 18 \\
Event-specific experiences ($|\mathcal{E}|$) & 3 \\
Calendar span & 1991--2026 (age 35) \\
High-resolution recency window & 5 years \\
High-resolution events in window & 3 \\
Average turns per high-resolution episode & 10 \\
Unique supporting characters & 16 \\
Generation time for this run & $\sim$8 minutes \\
\bottomrule
\end{tabular}}
\caption{Statistics for the generated $\mathcal{M}_\pi$ in the scientist example.}
\label{tab:memory-stats}
\end{table}

\section{PersonaGym Subset Sampling}
\label{app:subset_sampling}

The full PersonaGym benchmark contains 200 personas and 50 questions per persona, for 10{,}000 evaluation items. For ablation studies only, we use a 50-persona subset with 2{,}500 questions to reduce evaluation cost while preserving the major demographic and identity variations used in the full benchmark.

We construct the subset from the original persona descriptions, without using model outputs. Each persona is assigned to strata based on three attributes used in the ablation sampling protocol: age group, social background, and special identity. If an attribute is not explicit in the description, we assign it to an \texttt{unspecified} bin rather than inventing a label.

The subset construction proceeds as follows. We first reserve slots for rare or explicitly marked special-identity strata when such personas are present, then allocate the remaining slots approximately in proportion to the full benchmark distribution over age group and social background. Within each stratum, we sample uniformly from eligible personas and fill any remaining slots with personas that reduce the largest remaining stratum imbalance. Finally, we evaluate all 50 questions for each selected persona, yielding 2{,}500 ablation items.

This subset is used for the backbone and component ablations reported in the paper. Main PersonaGym results are reported on the full 200-persona benchmark unless explicitly stated otherwise.

\section{Normalised Factor Score}
\label{app:ss}

This section gives the exact attribution calculation used for the backbone ablation, so that the Normalised Factor Score in Figure~\ref{fig:ab-model-sscore} can be reproduced from the $2\times2$ cell means. We use a two-factor Shapley-style analysis to quantify how much PersonaScore variation is associated with the role-playing backbone and the memory-synthesis backbone. Let $C$ denote the role-playing model used by the final persona agent, and let $S$ denote the simulation model used to synthesize the memory base. In the $2\times2$ factorial ablation, condition $(c_i,s_j)$ yields mean PersonaScore $\mu_{ij}$.

We define the characteristic value function using the population variance of the corresponding cell means:
\begin{align*}
v(\emptyset) &= 0, \\
v(\{C\}) &= \frac{1}{2}\sum_{j=1}^{2}\operatorname{Var}_{i\in\{1,2\}}(\mu_{ij}), \\
v(\{S\}) &= \frac{1}{2}\sum_{i=1}^{2}\operatorname{Var}_{j\in\{1,2\}}(\mu_{ij}), \\
v(\{C,S\}) &= \operatorname{Var}_{(i,j)}(\mu_{ij}).
\end{align*}
The two-factor Shapley values are
\begin{align*}
\varphi_C &= \frac{1}{2}v(\{C\}) + \frac{1}{2}\left[v(\{C,S\}) - v(\{S\})\right], \\
\varphi_S &= \frac{1}{2}v(\{S\}) + \frac{1}{2}\left[v(\{C,S\}) - v(\{C\})\right].
\end{align*}
We then report the normalized attribution scores:
\begin{equation*}
\hat{\varphi}_C = \frac{\varphi_C}{\varphi_C+\varphi_S},\qquad
\hat{\varphi}_S = \frac{\varphi_S}{\varphi_C+\varphi_S}.
\end{equation*}
Applying this computation to our ablation gives $\hat{\varphi}_C=87.7\%$ and $\hat{\varphi}_S=12.3\%$, indicating that PersonaScore variation is dominated by the inference-time role-playing backbone rather than by the model used to synthesize the memory.

\end{document}